\documentclass{article}

\usepackage[final, eandd]{neurips_2026}

\usepackage[utf8]{inputenc}
\usepackage[T1]{fontenc}
\usepackage[implicit=false]{hyperref}
\hypersetup{colorlinks=true, allcolors=agBold}
\usepackage{url}
\usepackage{booktabs}
\usepackage{amsfonts}
\usepackage{amsmath}
\usepackage{nicefrac}
\usepackage{microtype}
\usepackage{xcolor}
\usepackage{graphicx}
\usepackage{subcaption}
\usepackage{float}
\usepackage{enumitem}
\usepackage{verbatim}
\usepackage{multirow}
\usepackage{wrapfig}
\usepackage{colortbl}
\usepackage{tikz}
\usepackage{eso-pic}

\definecolor{agBlue}{HTML}{60A5FA}
\definecolor{agNavy}{HTML}{1E3A5F}
\definecolor{agBgLight}{HTML}{F0F4FF}
\definecolor{agBgRow}{HTML}{E8EEFF}
\definecolor{agGrid}{HTML}{B8D4FF}
\definecolor{agHighlight}{HTML}{DBEAFE}

\newcommand{\agheader}{\rowcolor{agBlue}\color{agNavy}\bfseries}
\newcommand{\agrowA}{\rowcolor{agBgLight}}
\newcommand{\agrowB}{\rowcolor{white}}

\definecolor{agBold}{HTML}{3B82F6}
\newcommand{\revision}[1]{#1}

\newcommand{\agsprite}[1]{\raisebox{-0.2em}{\includegraphics[height=1.1em]{figures/#1}}\hspace{0.3em}}

\newcommand{\affillogo}[1]{%
  \IfFileExists{figures/#1}{%
    \raisebox{-0.15em}{\includegraphics[height=0.95em]{figures/#1}}\hspace{0.25em}%
  }{}%
}

\makeatletter
\newcommand\blfootnote[1]{%
  \begingroup
  \renewcommand\thefootnote{}%
  \footnotetext{#1}%
  \endgroup
}
\makeatother

\newcommand{\agbloglink}{https://roger-creus.github.io/agentick/blog/}
\newcommand{\agdocslink}{https://roger-creus.github.io/agentick/}
\newcommand{\agboardlink}{https://roger-creus.github.io/agentick/board/}
\newcommand{\agcodelink}{https://github.com/roger-creus/agentick}
\newcommand{\agdatasetslink}{https://huggingface.co/rogercc}

\title{Agentick: A Unified Benchmark for General Sequential Decision-Making Agents}

\author{%
  Roger Creus Castanyer~\affillogo{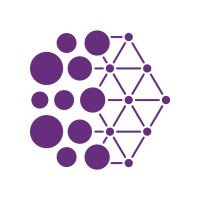}\affillogo{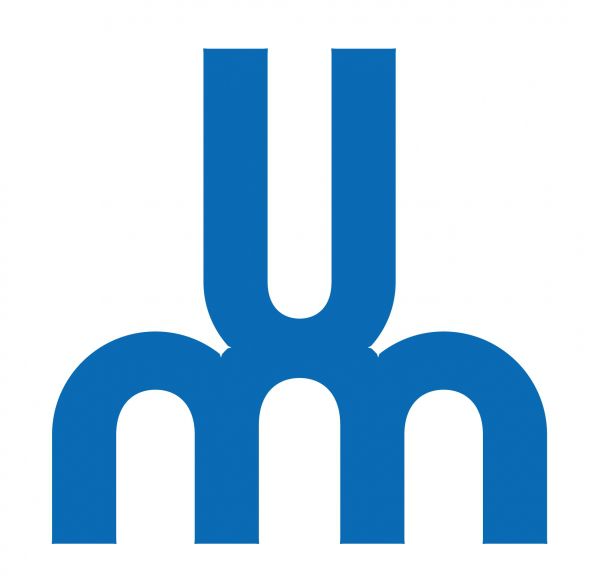} \\
  Mila Quebec AI Institute \\
  Universit\'e de Montr\'eal
  \And
  Pablo Samuel Castro$^{\ast}$~\affillogo{logo_mila.jpg}\affillogo{logo_udem.jpg}\affillogo{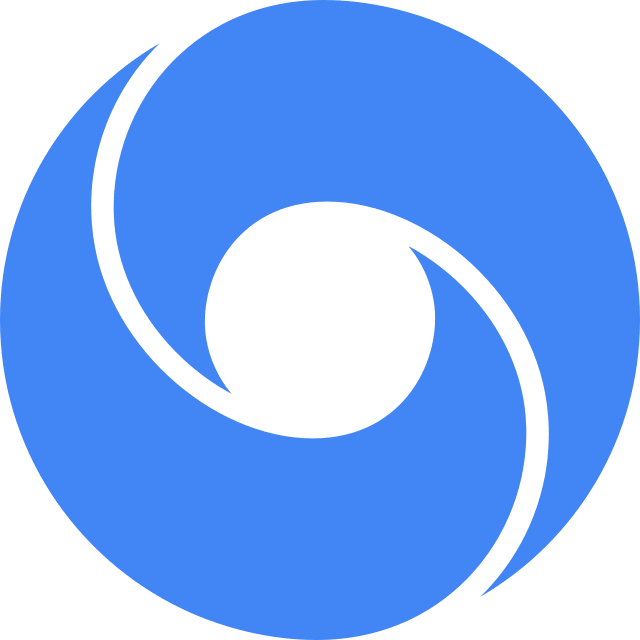} \\
  Mila Quebec AI Institute \\
  Universit\'e de Montr\'eal \\
  Google DeepMind
  \And
  Glen Berseth$^{\ast}$~\affillogo{logo_mila.jpg}\affillogo{logo_udem.jpg} \\
  Mila Quebec AI Institute \\
  Universit\'e de Montr\'eal
}

\AddToShipoutPictureBG{%
  \AtPageLowerLeft{%
    \raisebox{0.4cm}{\makebox[\paperwidth]{\includegraphics[width=0.15\paperwidth]{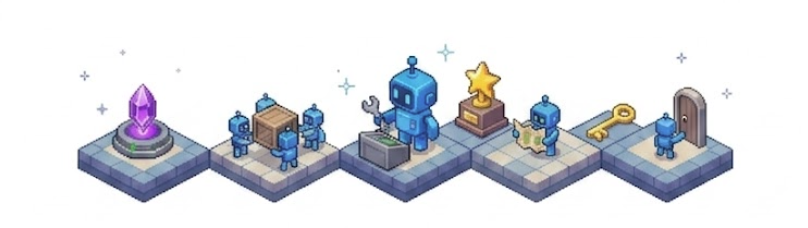}}}%
  }%
}

\begin{document}

\vspace*{-1.8cm}
\begin{center}
    \includegraphics[width=0.65\linewidth]{figures/banner_new.png}
\end{center}
\vspace{-1.2em}

\maketitle
\blfootnote{$^{\ast}$Equal supervision.}
\blfootnote{Correspondence to: \texttt{roger.creus-castanyer@mila.quebec}}

\begin{abstract}
AI agent research spans a wide spectrum: from RL agents that learn from scratch to foundation model agents that leverage pre-trained knowledge, \revision{yet no unified benchmark aligns task semantics, procedural instances, action semantics, and metrics across these approaches.} We present \textbf{Agentick}, a benchmark for sequential decision-making agents designed to evaluate RL, LLM, VLM, hybrid, and human agents on common ground and to power research on the fundamental challenges of sequential decision-making. Agentick provides 37 procedurally generated tasks across six capability categories, four difficulty levels, and five observation modalities, all exposed through a single Gymnasium-compatible interface. The benchmark ships with a Coding API, oracle reference policies for all tasks, pre-built SFT datasets, a composable agent harness, and a live leaderboard. \revision{Across 31 configurations and 108{,}550 episodes, SFT-250k leads overall at 0.447 ONS, GPT-5~mini leads generalization at 0.437, and PPO leads multi-agent at 0.432; the reasoning harness improves frozen LLMs by 3--10$\times$, and ASCII consistently outperforms natural language.}
These findings highlight the substantial room for improvement that remains across all agent paradigms. Agentick's capability-decomposed, multi-modal design provides the empirical infrastructure needed to drive progress toward general autonomous agents, both as an evaluation framework and as a training ground for RL post-training of foundation models in truly sequential environments.
\end{abstract}

\vspace{0.15em}
\begin{center}
\renewcommand{\arraystretch}{1.4}
\small
\begin{tabular}{@{}c@{\hspace{1.8em}}c@{\hspace{1.8em}}c@{\hspace{1.8em}}c@{\hspace{1.8em}}c@{}}
\agsprite{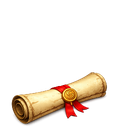}\,\href{\agbloglink}{\textbf{Blog}} &
\agsprite{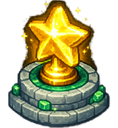}\,\href{\agdocslink}{\textbf{Docs}} &
\agsprite{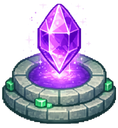}\,\href{\agboardlink}{\textbf{Leaderboard}} &
\agsprite{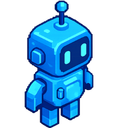}\,\href{\agcodelink}{\textbf{Code}} &
\agsprite{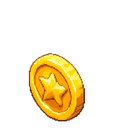}\,\href{\agdatasetslink}{\textbf{Datasets}}
\end{tabular}
\end{center}

\section{\agsprite{sprite_agent.png}Introduction}
\label{sec:introduction}

The pursuit of autonomous agents (systems that perceive their environment, reason about it, and take actions to achieve goals) has been a central objective of artificial intelligence research for decades~\citep{sutton1998reinforcement}.
The landscape of agent research now spans a wide spectrum of paradigms. At one end, deep reinforcement learning (RL) agents learn from scratch through environment interaction: PPO~\citep{schulman2017proximal}, DQN~\citep{mnih2015humanlevel}, and SAC~\citep{haarnoja2018soft} have achieved superhuman performance in Atari~\citep{bellemare2013arcade}, continuous control~\citep{tassa2018deepmind}, and strategic games~\citep{vinyals2019grandmaster}.
At the other end, foundation model (FM) agents, including large language models (LLMs) and vision-language models (VLMs) pre-trained on internet-scale data, leverage broad world knowledge for zero-shot decision-making through prompt engineering and inference-time scaling~\citep{yao2023react, wang2023voyager, ahn2022can}. Between these extremes lies a rich design space of hybrid approaches: FM-guided reward shaping~\citep{ma2023eureka, klissarov2023motif, castanyer2025arm}, RL post-training of foundation models~\citep{guo2025deepseek}, and FM-based skill discovery~\citep{klissarov2024maestromotif} and curriculum generation~\citep{wang2023voyager}.

Each paradigm makes different tradeoffs.
RL agents learn fine-grained control policies but are sample-inefficient, task-specific, and suffer from unstable optimization at scale~\citep{ceron2024value, castanyer2025stable, lyle2022understanding}. FM agents bring rich priors and semantic understanding but were not trained for control and struggle with precise, temporally extended actions~\citep{paglieri2024balrog}. This raises a central question: \textit{what combination of learning from interaction and pre-trained knowledge is needed to build fully capable autonomous agents?} Answering it requires the ability to compare agents across the full paradigm spectrum (RL from scratch, prompted foundation models, and hybrids in between) on the same tasks, and existing benchmarks cannot do this (Section~\ref{sec:related}).

\revision{We present \textbf{Agentick}, a benchmark designed from the ground up to align task semantics and evaluation protocols across the full agent design spectrum.} The design is guided by four principles: \revision{(1)~\textbf{paradigm-spanning interfaces} through five synchronized observation modalities suitable for different agent architectures;} (2)~\textbf{capability decomposition} across a variety of dimensions of sequential decision-making; (3)~\textbf{training-first design} with a programmatic Coding API, oracle policies, pre-built fine-tuning datasets, and vectorizable environments; and (4)~\textbf{controlled difficulty} with four levels per task and procedural generation for reproducibility. Agentick provides 37 tasks across navigation, planning, reasoning, memory, generalization, and multi-agent coordination, all through a standard interface~\citep{towers2024gymnasium}.

\revision{To validate the benchmark's discriminative power, we evaluate 31 configurations spanning frozen frontier and open-weight LLMs, pixel-based PPO trained from scratch, and Qwen3.5-4B fine-tuned on oracle trajectories. Three findings emerge. First, capability leaders remain complementary: SFT-250k leads overall (0.447 ONS) and four categories, GPT-5~mini leads generalization (0.437), and PPO leads multi-agent (0.432). Second, the reasoning harness improves frozen LLM performance by 3--10$\times$. Third, ASCII consistently outperforms natural language for LLM agents. These controlled comparisons are uniquely enabled by Agentick's shared tasks, seeds, metrics, and capability decomposition.}

The benchmark, code, documentation, pre-built datasets, and live leaderboard are publicly available (links above the abstract). Section~\ref{sec:related} positions Agentick relative to existing benchmarks. Section~\ref{sec:benchmark} describes the benchmark design. Section~\ref{sec:experiments} presents experimental results. Section~\ref{sec:discussion} discusses future directions and conclusions.

\section{\agsprite{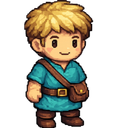}Related Work}
\label{sec:related}

Agent evaluation frameworks can be broadly grouped by the paradigm they target. Table~\ref{tab:comparison} provides a structured comparison; we discuss each group below.

\noindent \textbf{RL benchmarks.} The Arcade Learning Environment (ALE)~\citep{bellemare2013arcade} established the dominant evaluation paradigm for deep RL through Atari 2600 games with pixel observations, and the DeepMind Control Suite~\citep{tassa2018deepmind} extended this to continuous control with proprioceptive and pixel observations. bsuite~\citep{osband2020bsuite} took a diagnostic approach, designing experiments that isolate specific RL capabilities such as exploration, credit assignment, memory, and generalization in deliberately simple settings. MiniGrid~\citep{chevalier-boisvert2023minigrid} provides a modular gridworld framework for goal-oriented tasks. The NetHack Learning Environment~\citep{kuettler2020nethack} exposes a procedurally generated roguelike game with extreme partial observability and long horizons, representing one of the most challenging single-environment RL benchmarks, and MiniHack~\citep{samvelyan2021minihack} builds on it with a flexible domain-specific language for constructing diverse Gymnasium-compatible RL tasks. Crafter~\citep{hafner2022benchmarking} and its JAX-accelerated extension Craftax~\citep{matthews2024craftax} offer single procedurally generated survival games that test a broad spectrum of capabilities. Procgen~\citep{cobbe2020leveraging} provides procedurally generated game levels for studying generalization. These benchmarks remain extensively used for RL research, but were designed primarily for RL agents. Agentick is closest in spirit to MiniGrid and MiniHack, but differs by providing purpose-built capability categories, five synchronized observation modalities, standardized LLM/VLM/RL harnesses, oracle trajectory datasets, and a unified scoring protocol for cross-paradigm comparison.

\noindent \textbf{LLM and VLM agent benchmarks.} BALROG~\citep{paglieri2024balrog} wraps six existing RL game environments (BabyAI, Crafter, TextWorld, Baba~Is~AI, MiniHack, NetHack) into text and vision interfaces for LLM and VLM evaluation, demonstrating that even frontier models struggle at long-horizon interactive tasks. However, BALROG introduces no new tasks designed to target specific agentic capabilities, provides no capability decomposition or unified scoring across its heterogeneous game suite, and does not systematically investigate how observation modality affects agent performance. TextWorld~\citep{cote2019textworld} provides text-based adventure games for language grounding but targets text-only agents. Agentick builds on the key insight from BALROG that interactive tasks expose fundamental weaknesses in FM agents, while addressing these limitations through purpose-built tasks, multi-modal observations, and cross-paradigm evaluation infrastructure.

\noindent \textbf{RLVR training environments.} A parallel line of work uses verifiable environments for RL post-training of language models. Mathematical reasoning benchmarks such as MATH~\citep{hendrycks2021math} and GSM8K~\citep{cobbe2021gsm8k}, code generation benchmarks like SWE-bench~\citep{jimenez2024swebench} and HumanEval~\citep{chen2021evaluating}, and reasoning environments like Reasoning Gym~\citep{reasoninggym2025} are widely used for RLVR methods~\citep{guo2025deepseek}. These environments are valuable for eliciting reasoning capabilities, but they operate in settings with limited sequential complexity: episodes are single-turn or short-horizon, transitions are fully deterministic, and there is no partial observability, stochastic dynamics, or multi-agent interaction. Agentick also provides verifiable rewards, but is designed from the ground up to test the fundamental challenges that emerge in sequential decision-making: \revision{interactive, stochastic, multi-step environments (configured horizons 20--800) with} partial observability, exploration, multi-step credit assignment, and multi-agent coordination: the kinds of challenges that current RLVR benchmarks do not expose.

\noindent \textbf{Interactive reasoning benchmarks.} ARC-AGI-3~\citep{chollet2026arcagi3} introduces 135 interactive turn-based grid environments for agentic evaluation, supporting both RL and LLM agents. However, it evaluates through a single aggregate score without capability decomposition, provides limited public environments, and uses a custom SDK rather than a standard RL interface such as Gymnasium. Agentick occupies a different point in the design space: it is built to be both useful for evaluating frontier model capabilities and accessible for academic research, with Gymnasium-native environments, procedural generation for training, multi-modal observations, and per-category diagnostic scoring.

\begin{table}[t]
  \caption{Comparison of agent evaluation frameworks. Agentick is the only benchmark supporting all agent paradigms with capability decomposition and training infrastructure.}
  \label{tab:comparison}
  \centering
  \footnotesize
  \setlength{\tabcolsep}{10pt}
  \begin{tabular}{lccccccc}
    \agheader & \#Tasks & RL & LLM & Obs.\ Modes & Cap.\ Dec. & Train Data & Gym \\
    \agrowA ALE & 57 & \checkmark & \texttimes & 1 & \texttimes & \texttimes & \checkmark \\
    \agrowB DM Control & 30 & \checkmark & \texttimes & 2 & \texttimes & \texttimes & \texttimes \\
    \agrowA bsuite & 23 & \checkmark & \texttimes & 1 & \checkmark & \texttimes & \texttimes \\
    \agrowB MiniGrid & 20+ & \checkmark & \texttimes & 2 & \texttimes & \texttimes & \checkmark \\
    \agrowA MiniHack & 100+ & \checkmark & \texttimes & 2+ & \texttimes & \texttimes & \checkmark \\
    \agrowB Crafter & 1 & \checkmark & \texttimes & 1 & \texttimes & \texttimes & \checkmark \\
    \agrowA Craftax & 2 & \checkmark & \texttimes & 2 & \texttimes & \texttimes & \checkmark \\
    \agrowB NetHack & 1 & \checkmark & \texttimes & 2 & \texttimes & \texttimes & \checkmark \\
    \agrowA BALROG & 6 & \texttimes & \checkmark & 2 & \texttimes & \texttimes & \checkmark \\
    \agrowB ARC-AGI-3 & 135 & \checkmark & \checkmark & 1 & \texttimes & \texttimes & \texttimes \\
    \midrule
    \rowcolor{agHighlight} \textbf{Agentick} & \textbf{37} & \checkmark & \checkmark & \textbf{5} & \checkmark & \checkmark & \checkmark \\
  \end{tabular}
\end{table}

\section{\agsprite{sprite_goal.png}The Agentick Benchmark}
\label{sec:benchmark}

Agentick provides 37 procedurally generated gridworld tasks across six capability categories, five observation modalities, four difficulty levels, and a complete training and evaluation pipeline, all through a unified Gymnasium-compatible~\citep{towers2024gymnasium} interface.

\subsection{Design, Tasks, and Observations}
\label{sec:design}

The design is guided by four principles.

\begin{figure}[t]
    \centering
    \begin{minipage}[c]{0.30\linewidth}
        \centering
        \includegraphics[width=\linewidth]{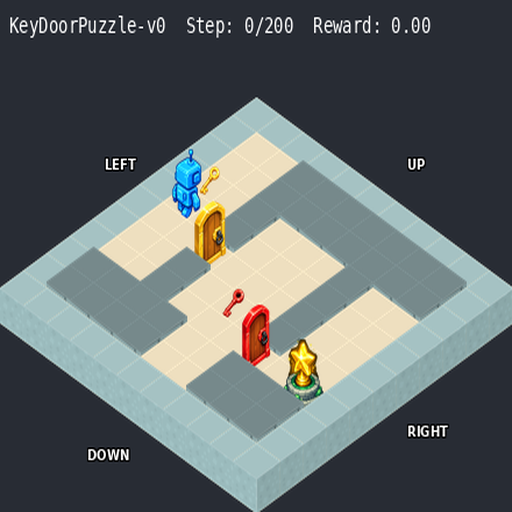}
    \end{minipage}
    \hfill
    \begin{minipage}[c]{0.65\linewidth}
        {\small
\begin{verbatim}
 # # # # # # # # # # #   Legend:
 # . . # # # # # # # #   ^ v < > : Agent
 # . . # # # # # # # #   # : Wall  . : Empty
 # Kg. # . . . # . . #   G : Goal  K : Key
 # ^ . # . . . # . . #   D : Door (closed)
 # . . Dg. . . # . . #   g=gold r=red
 # . . # . Kr. # . . #
 # . . # . . . Dr. G #
 # # # # # . . # # # #
 # # # # # . . # # # #
 # # # # # # # # # # #
\end{verbatim}
}
    \end{minipage}
    \vspace{-0.5em}
    \caption{Two observation modalities for KeyDoorPuzzle at medium difficulty. \textbf{Left:} isometric pixel rendering (512$\times$512 by default, resizable for RL preprocessing) for VLM agents, CNN-based RL, and human play. \textbf{Right:} ASCII grid for LLM agents; the agent (\texttt{\^{}}) must collect the gold key (\texttt{Kg}), open the gold door (\texttt{Dg}), then the red key (\texttt{Kr}) and red door (\texttt{Dr}) to reach the goal (\texttt{G}). The same state also produces natural language, structured dictionary, and numpy array observations.}
    \label{fig:obs_example}
\end{figure}

\revision{\textbf{Paradigm-spanning interfaces}:} every task produces five observation modalities simultaneously (ASCII text grids, natural language descriptions, structured dictionaries, 512$\times$512 isometric pixel renderings, and raw numpy state arrays), \revision{so RL, LLM, VLM, and human agents can use interfaces suited to their architectures while sharing the same underlying environment state, transition, and action semantics} (Figure~\ref{fig:obs_example}; Appendix~\ref{app:obs_modalities} shows all five for the same state). Pixel observations are returned at 512$\times$512 by default for VLM and human use, but can be resized through standard wrappers; our PPO baselines use 84$\times$84 grayscale frame-stacked images following the ALE preprocessing convention, obtained by bilinearly resizing the native 512$\times$512 RGB renders and converting to luminance.

\textbf{Capability decomposition}: rather than a single aggregate score, evaluation is decomposed along six capability dimensions, enabling radar-chart profiling that reveals where an agent excels and where it falls short. 

\textbf{Training-first design}: environments are vectorizable for parallel RL training, oracle policies built on a programmatic Coding API (Appendix~\ref{app:coding_api}) are provided for all 37 tasks, and pre-built SFT datasets of 120K--500K oracle episodes are available on HuggingFace. 

\textbf{Controlled difficulty}: four levels per task (easy through expert) scale grid size, constraint complexity, object count, and episode length, with procedural generation ensuring unique layouts at every seed (see Appendix~\ref{app:difficulty} for visual examples). \revision{Difficulty labels select parameter regimes of each seeded generator, never hand-authored layouts; Table~\ref{tab:difficulty_validation} shows the resulting performance degradation by category for four representative agents.}

Beyond the fixed leaderboard protocol, Agentick is designed as a configurable experimental substrate. Researchers can select arbitrary task and difficulty subsets, choose any observation modality, resize or otherwise preprocess pixel observations through standard wrappers, and vary language or prompt templates through the harness interface while preserving the same environment seeds and scoring protocol. This enables controlled studies of modality, instruction design, curricula, and agent scaffolding without changing the underlying task semantics.

The 37 tasks are organized into six categories (Appendix~\ref{app:tasks} provides full descriptions and a visual gallery). \textbf{Navigation} (8 tasks) evaluates spatial reasoning, pathfinding, reactive control, and instruction grounding. \textbf{Planning} (9 tasks) requires multi-step lookahead, constraint satisfaction, backtracking, and resource allocation. \textbf{Reasoning} (8 tasks) demands logical inference, pattern matching, abstraction, and resistance to misleading rewards. \textbf{Memory} (4 tasks) tests information retention over extended horizons under partial observability. \textbf{Generalization} (3 tasks) evaluates few-shot rule inference, adaptation to shifting dynamics, and noise robustness. \textbf{Multi-Agent} (5 tasks) requires coordination with or against scripted agents with diverse behavioral profiles.

\subsection{Evaluation Framework}
\label{sec:evaluation}

The primary metric is the \textbf{Oracle-Normalized Score} (ONS):
\begin{equation}
    \text{ONS} = \frac{\text{agent\_return} - \text{random\_baseline}}{\text{oracle\_return} - \text{random\_baseline}}
    \label{eq:ons}
\end{equation}
where 0.0 corresponds to random agent performance and 1.0 to the oracle upper bound. Scores above 1.0 are possible when an agent outperforms the oracle reference policy. ONS is computed per (task, difficulty) pair and aggregated per-category and overall via the arithmetic mean. This normalization accounts for differences in task difficulty and reward scale, enabling meaningful cross-task comparison, analogous to human-normalized scores in ALE~\citep{mnih2015humanlevel} but calibrated to task-specific reference policies rather than human play.

Evaluation uses 25 deterministic seeds per task-difficulty pair (37~tasks $\times$ 4~difficulties $\times$ 25~seeds $=$ 3{,}700 total episodes per agent), derived from SHA-256 hashes of \texttt{"\{task\}::\{difficulty\}::eval"} to ensure every submission runs on identical episodes. A separate pool of 2,000 training seeds per task-difficulty is available for within-benchmark learning. We report 95\% bootstrap confidence intervals~\citep{agarwal2021deep} computed over the 25 evaluation episodes per task-difficulty pair, and provide a YAML-based experiment runner with parallel execution and standardized output.

The oracle upper bound in ONS is computed from oracle reference policies\footnote{Developed with the aid of coding agents and iterative refinement.} implemented through Agentick's \textbf{Coding API} (\texttt{AgentickAPI}), a programmatic interface that exposes spatial queries, BFS pathfinding, entity lookups, and high-level action primitives (full documentation in Appendix~\ref{app:coding_api}). Using this API, we built sophisticated programmatic solvers for all 37 tasks that achieve near-optimal performance in most cases. We use the term ``oracle'' to denote the strongest available reference policy for each task; some oracles fall short of true optimality on tasks with stochastic elements (e.g., scripted agents that wander unpredictably), where no deterministic policy can guarantee the best outcome on every seed, but they nevertheless achieve high success rates and serve as strong upper bounds for ONS calibration.

\subsection{Training Infrastructure and Agent Harnesses}
\label{sec:infrastructure}

\noindent \textbf{Oracle datasets.} The oracle reference policies described above also serve as expert trajectory generators. Pre-built datasets of 120K, 250K, and 500K oracle episodes are available on HuggingFace,\footnote{Croissant metadata files (including the 2026 Responsible-AI fields) for all three datasets are bundled in the supplementary repository.} containing per-step ASCII and language observations, actions, rewards, and \textit{done} flags across all tasks and difficulties. These datasets enable supervised fine-tuning (SFT) and behavior cloning within the benchmark. The Coding API itself is also available for hand-coded bots, planners, and code-generating LLM agents.

\noindent \textbf{Composable agent harness.} Agentick provides a modular harness infrastructure for LLM and VLM agents. A \texttt{BaseAgent} composes a \texttt{ModelBackend} (e.g. from OpenAI, Gemini, HuggingFace or vLLM providers) with a \texttt{HarnessPreset} that controls the full inference-time strategy: the complete pipeline that transforms environment observations into actions at each step, including the system prompt, observation formatting, history management, reasoning elicitation, and action parsing. Researchers can implement custom harness presets by subclassing a simple interface, enabling systematic study of how harness design affects agent performance. Two built-in presets serve as baselines: \textit{Markovian} (receives only the current observation and outputs a single action integer, with no intermediate reasoning) and \textit{Markovian Reasoner} (also receives only the current observation, but prompts the model for concise chain-of-thought before action selection). Both are memoryless and support any observation mode. By explicitly decoupling inference-time strategy from the underlying model, the harness abstraction makes it straightforward to evaluate the same model under different prompting designs or to develop learned harnesses that optimize strategies through experience. Experiments are configured via YAML files specifying model, backend, harness, observation mode, tasks, difficulties, and seed counts.

\noindent \textbf{Leaderboard.} A public leaderboard\footnote{\url{\agboardlink}} enables standardized comparison. Each submission records per-task ONS, per-category aggregates, and overall ONS, alongside metadata such as model family, parameter count, harness preset, and observation mode. The leaderboard displays radar-chart capability profiles (as in Figure~\ref{fig:radar}) that visually decompose each agent's strengths and weaknesses across the six capability categories. All evaluation code and seed definitions are public, so any result can be independently reproduced.

\section{\agsprite{sprite_gem.png}Experiments}
\label{sec:experiments}

We evaluate agents spanning the full paradigm spectrum to validate that Agentick produces discriminative, actionable insights. The goal is not to crown a ``best agent'' but to demonstrate that the benchmark reveals meaningful differences across paradigms, capabilities, observation modes, and prompting strategies.

\subsection{Experimental Setup}
\label{sec:setup}

\revision{We report 31 agent and reference configurations under the shared evaluation protocol, totaling 108{,}550 completed episodes:}

\begin{itemize}[leftmargin=*, itemsep=1pt, parsep=0pt]
    \item \textbf{Frontier LLMs}\footnote{These models belong to the latest release families (GPT-5, Gemini~3.1, Claude~4.5) but are the more economical variants (mini, Flash~Lite, Haiku) rather than the flagship models (Pro, Opus). Budget constraints prevented evaluation of the full-scale frontier models at the time of writing; we look forward to populating the leaderboard with those results.}: GPT-5~mini, Gemini~3.1~Flash~Lite, and Claude~Haiku~4.5, evaluated with ASCII observations and the chain-of-thought Markovian Reasoner harness.
    \item \textbf{RL from scratch}: PPO~\citep{schulman2017proximal} trained with Stable-Baselines3~\citep{raffin2021stable} from pixel observations using the standard ALE-style preprocessing pipeline: 512$\times$512 isometric renders resized to 84$\times$84, converted to grayscale, and stacked over four frames. We report three configurations: \textit{PPO Dense (2M)}, the headline RL baseline, trained for 2M steps with dense reward shaping; \textit{PPO Dense (500k)}, a shorter-budget control trained for 500k steps with the same dense rewards; and \textit{PPO Sparse (500k)}, a 500k-step run with terminal-only sparse rewards, included to probe the sensitivity of the RL signal to reward shaping under limited compute.
    \item \textbf{Open-weight LLMs}: Qwen3-4B and Qwen3.5 at 0.8B, 2B, and 4B parameters~\citep{qwen3technicalreport}, evaluated across observation modes and harnesses. \revision{We additionally evaluate Qwen3.5-4B after LoRA SFT on 120k or 250k oracle state--action examples.}
\end{itemize}

All agents are evaluated on the official deterministic evaluation seeds. We report 95\% bootstrap confidence intervals~\citep{agarwal2021deep} computed over the 25 evaluation episodes per task-difficulty pair; error bars are shown in all figures.

\paragraph{\revision{PPO budget and learning dynamics.}}
\revision{PPO is single-task: one independent policy per task--difficulty pair (148 policies; 296M aggregate steps). Final 25-seed evaluation from 500k to 2M changes mean success from 0.135 to 0.196 and ONS from 0.2258 to 0.2869. For the 126 recovered periodic curves, task-balanced smoothed success at 500k/1M/1.5M/2M is 0.243/0.329/0.377/0.393: successive gains shrink from $+0.086$ to $+0.048$ to $+0.016$. From 1.5M to 2M, 32/37 task aggregates change by at most 0.05 and 36/37 by at most 0.10. Appendix Table~\ref{tab:ppo_curve_checkpoints} and Figures~\ref{fig:ppo_curves_navigation}--\ref{fig:ppo_curves_multiagent} provide the full evidence.}

\subsection{Overall Performance}
\label{sec:overall}

\begin{figure}[t]
    \centering
    \begin{minipage}[t]{0.48\textwidth}
        \centering
        \includegraphics[width=\linewidth]{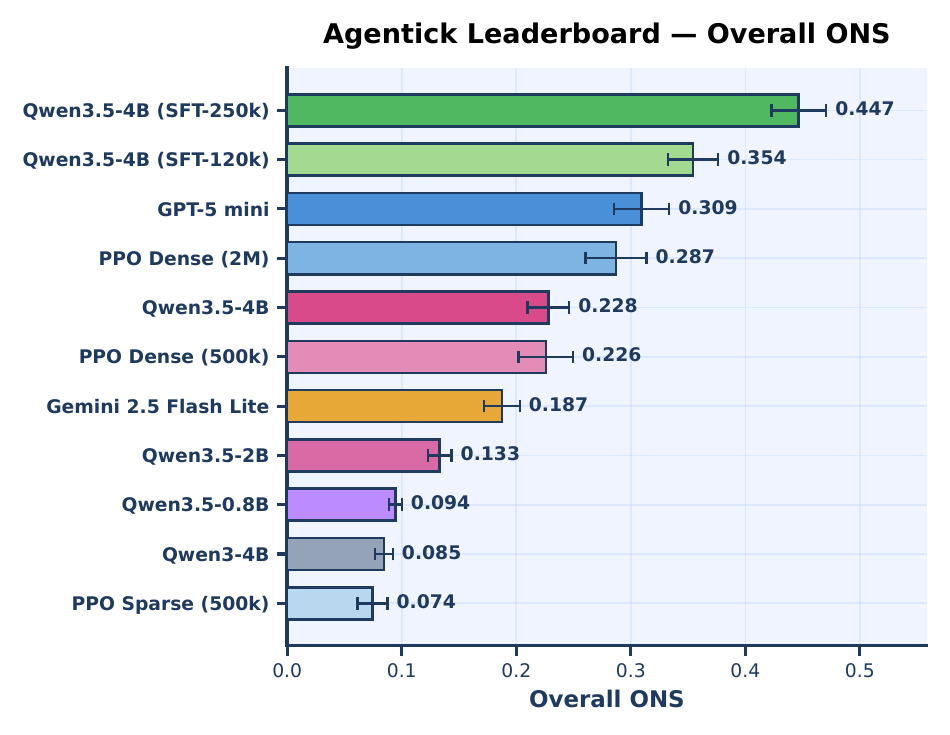}
        \caption{\revision{Overall ONS using the latest complete evaluations. SFT-250k leads at 0.447; frozen GPT-5~mini and PPO (2M) reach 0.309 and 0.287.}}
        \label{fig:overall_ons}
    \end{minipage}
    \hfill
    \begin{minipage}[t]{0.48\textwidth}
        \centering
        \includegraphics[width=\linewidth]{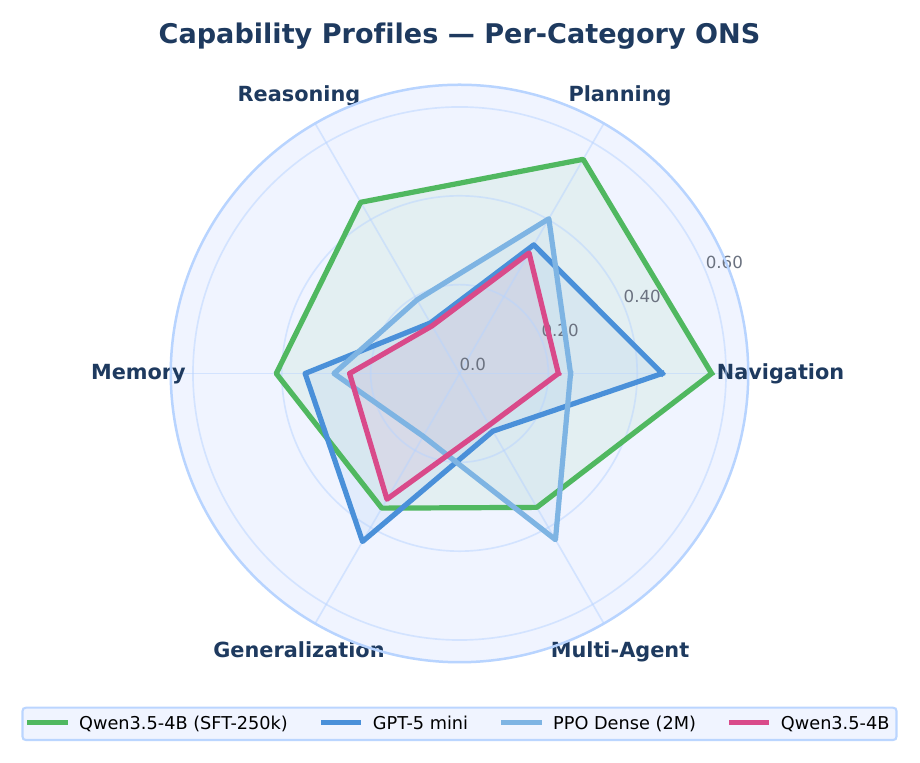}
        \caption{\revision{Category profiles for representative fine-tuned, frozen frontier, RL, and frozen open-weight agents. Their strengths remain complementary.}}
        \label{fig:radar}
    \end{minipage}
\end{figure}

\revision{Figure~\ref{fig:overall_ons} shows the latest complete rankings. SFT-250k leads at 0.447 ONS, followed by SFT-120k (0.354), frozen GPT-5~mini (0.309), pixel PPO trained from scratch (0.287), and frozen Qwen3.5-4B with the Reasoner harness (0.228). Even the leader remains below half the oracle ceiling. These are evaluations under shared tasks, seeds, and metrics, not compute-matched training comparisons: frozen LLMs use zero Agentick training interactions, PPO learns through environment interaction, and SFT uses oracle trajectories. Reporting each regime is precisely what exposes their different transfer and learning profiles.}

\subsection{Capability Decomposition}
\label{sec:capability}

\begin{figure}[h!]
    \centering
    \includegraphics[width=0.85\linewidth]{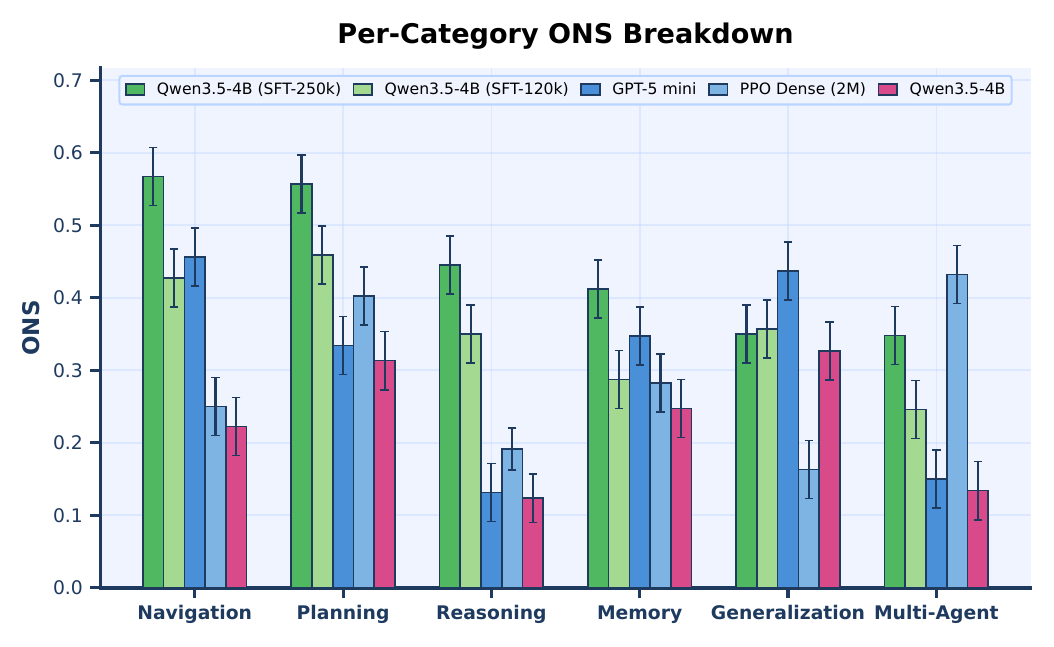}
    \vspace{-0.5em}
    \caption{\revision{Per-category ONS for the five highest overall configurations. SFT-250k leads four categories, GPT-5~mini leads generalization, and PPO leads multi-agent.}}
    \label{fig:category_breakdown}
\end{figure}

The per-category breakdown (Figures~\ref{fig:radar} and~\ref{fig:category_breakdown}) reveals a key finding: \textbf{the best agent varies by capability dimension}.

\revision{SFT-250k leads navigation (0.568), planning (0.557), reasoning (0.445), and memory (0.413), while frozen GPT-5~mini leads generalization (0.437) and PPO leads multi-agent (0.432). Thus fine-tuning substantially improves the base Qwen3.5-4B policy, yet does not erase the distinct transfer advantage of a frozen frontier model or PPO's interaction-trained coordination advantage.}

\subsection{Frontier LLMs on Hard Tasks}
\label{sec:frontier}

\begin{figure}[h!]
    \centering
    \includegraphics[width=\linewidth]{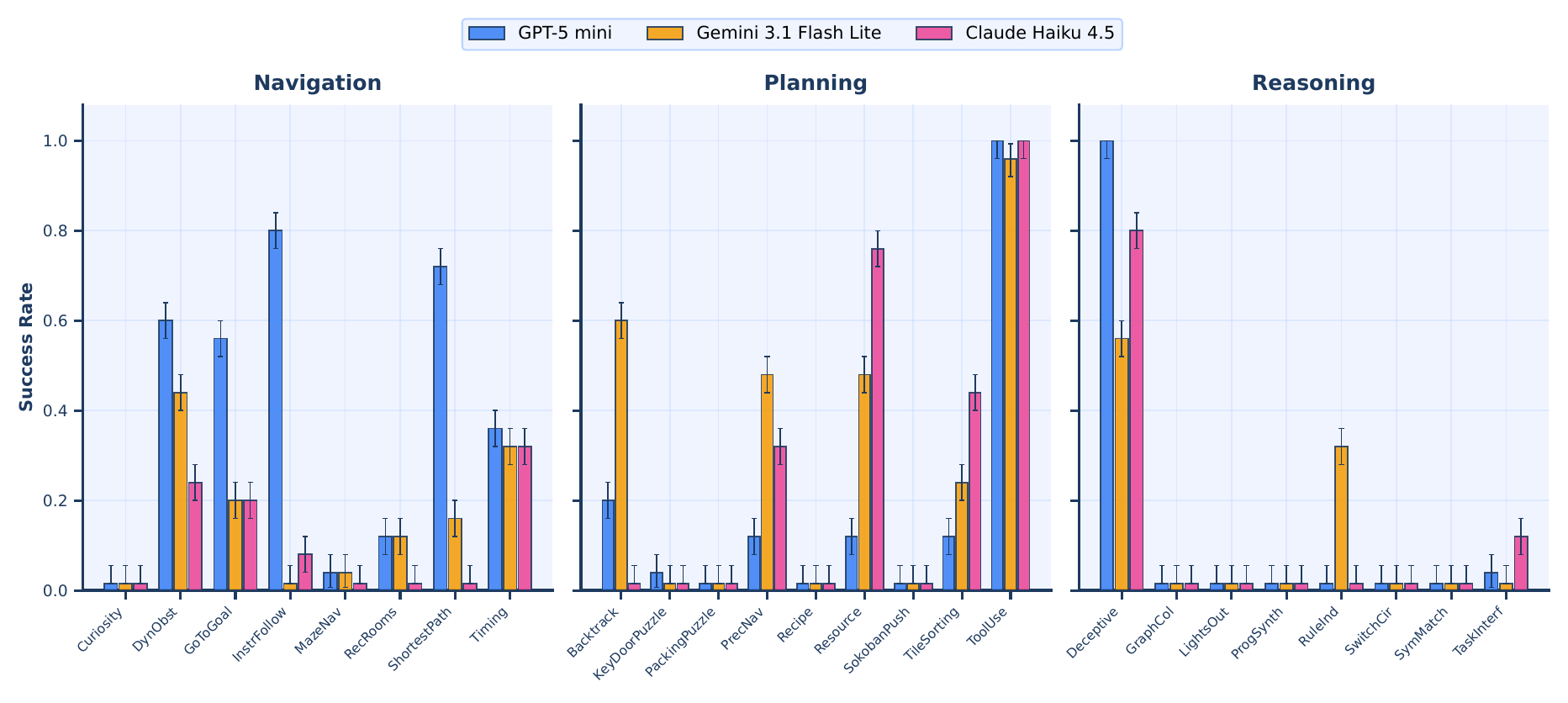}
    \caption{Per-task success rates at hard difficulty for three frontier LLMs. No single model dominates. Reasoning tasks (right panel) yield near-zero success for all models, exposing a fundamental limitation of prompting-only approaches.}
    \label{fig:frontier_hard}
\end{figure}

Figure~\ref{fig:frontier_hard} shows per-task success rates at hard difficulty for three frontier LLMs. No single model dominates across categories: GPT-5~mini leads most navigation tasks, but Haiku~4.5 outperforms on ResourceManagement (0.76 vs.\ 0.12), and all three models solve ToolUse at 0.96--1.0, where compositional reasoning is rewarded. Critically, \textbf{reasoning tasks remain largely intractable}: GraphColoring, LightsOut, SwitchCircuit, ProgramSynthesis, and SymbolMatching yield 0.0 success for all three models, suggesting that systematic search and state tracking cannot be achieved through prompting alone.

\subsection{Harness Design Matters: Observation Mode and Prompting Strategy}
\label{sec:harness}

\begin{figure}[h]
    \centering
    \includegraphics[width=0.64\linewidth]{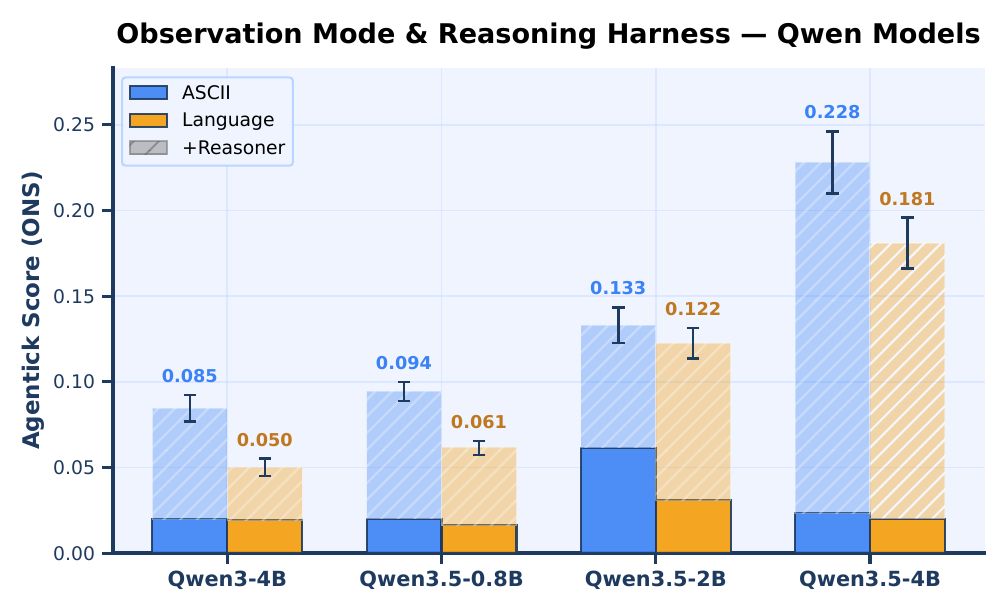}
    \caption{Qwen ONS by observation and harness. The Reasoner gain (lighter portion) is 3--10$\times$; ASCII (blue) consistently outperforms language (orange).}
    \label{fig:qwen_harness}
\end{figure}

A key motivation for Agentick's composable harness infrastructure is the hypothesis that \textit{how} an LLM agent is prompted matters as much as \textit{which} model is used. To test this, we evaluate the Qwen model family across all combinations of observation mode (ASCII vs.\ language) and harness preset (Markovian vs.\ Markovian Reasoner), using only the two built-in presets as a minimal illustration (Figure~\ref{fig:qwen_harness}).

Even with this simple comparison, the effect is striking. \textbf{The Reasoner harness multiplies performance by 3--10$\times$} across every model and observation mode: Qwen3.5-4B jumps from 0.023 to 0.228 ONS when switching from Markovian to Markovian Reasoner on ASCII observations, and even the smallest Qwen3.5-0.8B improves from 0.020 to 0.094. Additionally, \textbf{ASCII consistently outperforms language}: Qwen3.5-4B with the Reasoner achieves 0.228 on ASCII versus 0.181 on language, suggesting that compact grid representations are more effective for spatial reasoning than verbose natural language descriptions. It is also notable that Qwen3.5-0.8B (0.094) outperforms the much larger Qwen3-4B (0.085) despite having five times fewer parameters, illustrating how Agentick can surface generational improvements across model families, and how it will continue to be useful for benchmarking future generations of models as they are released.

These results establish harness design as a first-class variable, with the two built-in presets serving as an initial illustration. Agentick's composable architecture also supports non-Markovian, self-reflective, and learned harnesses, allowing the scaffolding around a model to be optimized rather than fixed.

\subsection{\revision{Training on Oracle Trajectories}}
\label{sec:sft}

\revision{To test whether Agentick supports adaptation rather than evaluation alone, we train rank-16 LoRA adapters for Qwen3.5-4B on either 120k or 250k per-step examples generated by the released oracle policies. Each example contains the current ASCII observation and oracle action only: no chain-of-thought rationale and no trajectory history. We select checkpoints after approximately two epochs, then evaluate them on the same 37 tasks, four difficulty levels, and 25 official seeds as the frozen model, under both the direct Markovian and Markovian Reasoner harnesses.}

\begin{figure}[t]
    \centering
    \includegraphics[width=\linewidth]{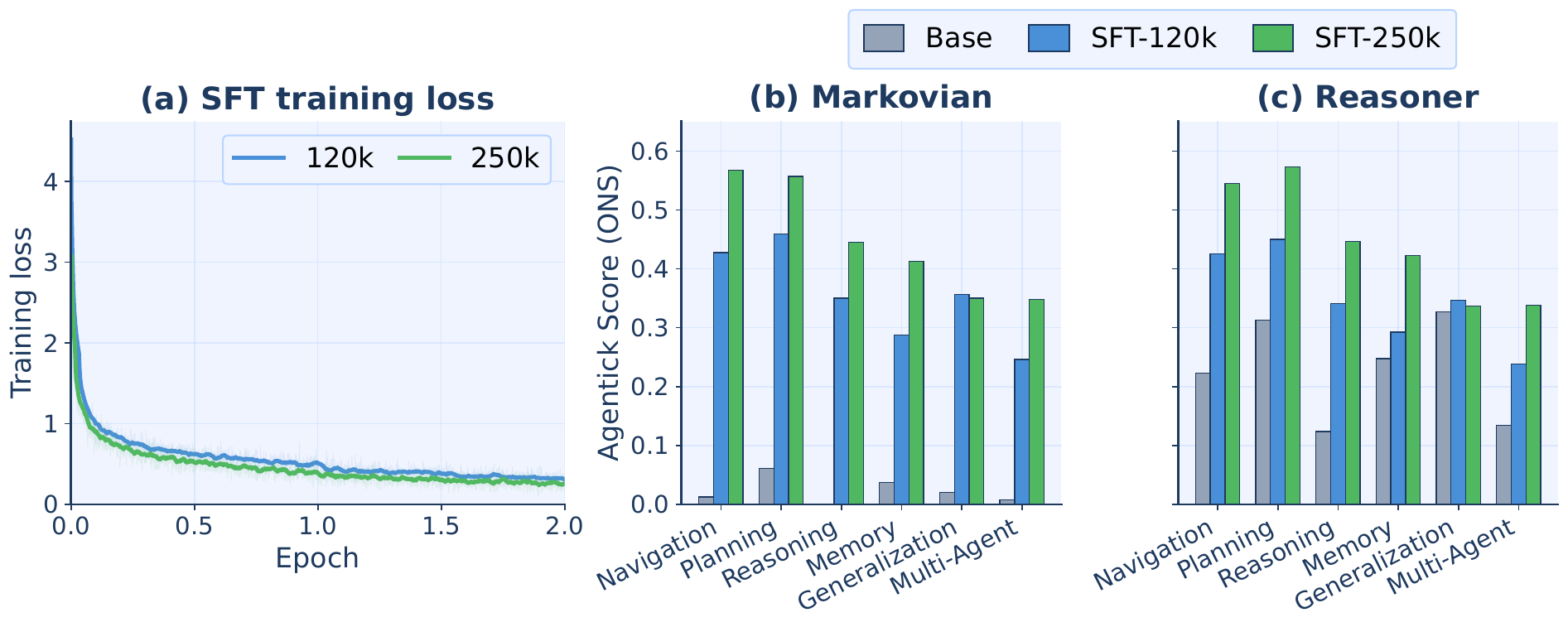}
    \caption{\revision{\textbf{SFT on Agentick oracle trajectories.} (a) Training loss for 120k- and 250k-example datasets. (b--c) Per-category evaluation for the frozen base model and two SFT checkpoints under direct Markovian and Reasoner harnesses. Overall ONS reaches 0.4466/0.4436 with 250k examples.}}
    \label{fig:sft_results}
\end{figure}

\revision{Figure~\ref{fig:sft_results} shows smooth optimization and large, data-scaled evaluation gains. Direct Markovian ONS rises from 0.0232 to 0.3544 with 120k examples and 0.4466 with 250k; Reasoner ONS rises from 0.2280 to 0.3489 and 0.4436. Thus, scaling 120k$\rightarrow$250k adds a further 0.0922/0.0947 ONS under the two harnesses. More importantly, the 0.2048 gap between the frozen direct and Reasoner policies collapses to 0.0030 after SFT. The direct SFT policy therefore no longer depends on prompted reasoning to act effectively, consistent with the demonstrations internalizing a state-to-action policy. SFT-250k leads navigation (0.568), planning (0.557), reasoning (0.445), and memory (0.413), while generalization (0.350) and multi-agent coordination (0.348) remain the clearest opportunities for methods beyond behavioral cloning.}

\section{\agsprite{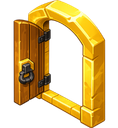}Discussion and Conclusion}
\label{sec:discussion}

\noindent \textbf{Future directions.} The most promising direction enabled by Agentick is \textbf{RL post-training of foundation models for sequential decision-making}. The progression from RLHF~\citep{ouyang2022training} to RLVR~\citep{guo2025deepseek} shows that RL can elicit capabilities from pretrained models, but current successes remain largely deterministic and single-turn. Extending this program to interactive, stochastic, multi-step environments introduces credit assignment, strategic exploration, and coordination~\citep{silver2025welcome}. Agentick supplies vectorizable environments, oracle demonstrations for warm starts, synchronized modalities, and capability-level diagnostics for this setting. Additional directions include VLM evaluation on isometric observations, learned or self-improving harnesses, curricula across procedural difficulty, and model-scaffolded task generation using Agentick's task primitives and verifiable rewards.

\noindent \textbf{Limitations.} Agentick is a controlled diagnostic suite: its discrete, 2D, turn-based tasks do not represent continuous-control, real-time, open-ended, or continual-learning settings. The six capability labels are primary, non-exclusive diagnostic groupings rather than a validated cognitive taxonomy, and the programmatic reference policies are not guaranteed globally optimal in stochastic tasks. \revision{We now evaluate frozen, RL-trained, and supervised fine-tuned agents, but not VLMs, hybrid RL/FM agents, or flagship frontier models.} Cross-paradigm profiles share the environment contract but are not compute-, supervision-, or modality-matched. Appendix~\ref{app:limitations} provides the full discussion.

\noindent \textbf{Broader impact.} Agentick is a research benchmark for controlled evaluation. It may accelerate progress on autonomous agents, with positive applications in assistive systems and scientific discovery as well as risks from agents acting without adequate oversight. We encourage safety and alignment considerations when developing agents in the benchmark.

\noindent \textbf{Conclusion.} Agentick provides 37 procedural task families, six diagnostic categories, five synchronized modalities, and training-first infrastructure under an aligned environment and evaluation contract. \revision{Across 31 configurations and 108{,}550 evaluation episodes, SFT-250k leads overall, GPT-5~mini retains the strongest generalization profile, PPO leads multi-agent coordination, and substantial gaps to the reference policies remain.} The benchmark, oracle datasets, evaluation harness, and leaderboard are public foundations for studying how training, representation, and inference-time strategy shape sequential decision-making.


\begin{ack}
We want to acknowledge funding support from Natural Sciences and Engineering Research Council (NSERC) of Canada, Fonds de recherche du Qu\'ebec (FRQNT; \url{https://doi.org/10.69777/2004792}) and The Canadian Institute for Advanced Research (CIFAR) and compute support from Digital Research Alliance of Canada, Mila IDT, and NVidia.
\end{ack}

\clearpage
\bibliographystyle{plainnat}
\bibliography{references}

\appendix
\clearpage
\section{Detailed Limitations}
\label{app:limitations}

We identify six limitations of the current version of Agentick:

\begin{enumerate}[leftmargin=*, itemsep=2pt]
    \item \textit{Gridworld abstraction.} All tasks are discrete, 2D, and turn-based. This is a deliberate choice (discrete control is the common denominator across all agent types) but it limits applicability to continuous control or real-time settings. \revision{The benchmark remains difficult despite this abstraction: the best agent reaches 0.447 ONS.}

    \item \textit{Capability coverage.} While 37 tasks across six categories is broad, the current set does not cover all fundamental challenges of sequential decision-making. Continual learning, open-ended language grounding, and long-horizon tool use in unstructured environments are not yet represented. The benchmark is designed to be actively developed and maintained, and we plan to add tasks targeting these capabilities over time.

    \item \textit{Observation equivalence.} Although all modalities expose the same underlying state, they are not informationally equivalent; for instance, isometric renders may occlude certain tiles due to perspective. The benchmark evaluates decision-making \textit{given} a particular observation format, not observation-agnostic intelligence.

    \item \textit{RL training budget.} \revision{Each task--difficulty policy was trained for 2M steps. The recovered curves are nearly stationary over their final 0.8M steps for 36/37 task aggregates, supporting this practical cap but not proving strict convergence; longer training, other algorithms, or curricula may still change the profile.}

    \item \textit{Multi-agent scope.} Opponent agents use built-in scripted policies rather than learned ones, limiting the study of emergent dynamics with co-adapting agents.

    \item \textit{Evaluation coverage.} \revision{Our results now cover frozen foundation models, single-task PPO from pixels, and Qwen3.5-4B fine-tuned on oracle trajectories, but not VLM agents, hybrid RL/FM agents, RL post-training of foundation models, or flagship frontier models.}
\end{enumerate}

\clearpage
\section{Difficulty Scaling}
\label{app:difficulty}

\begin{figure}[H]
    \centering
    \includegraphics[width=\linewidth]{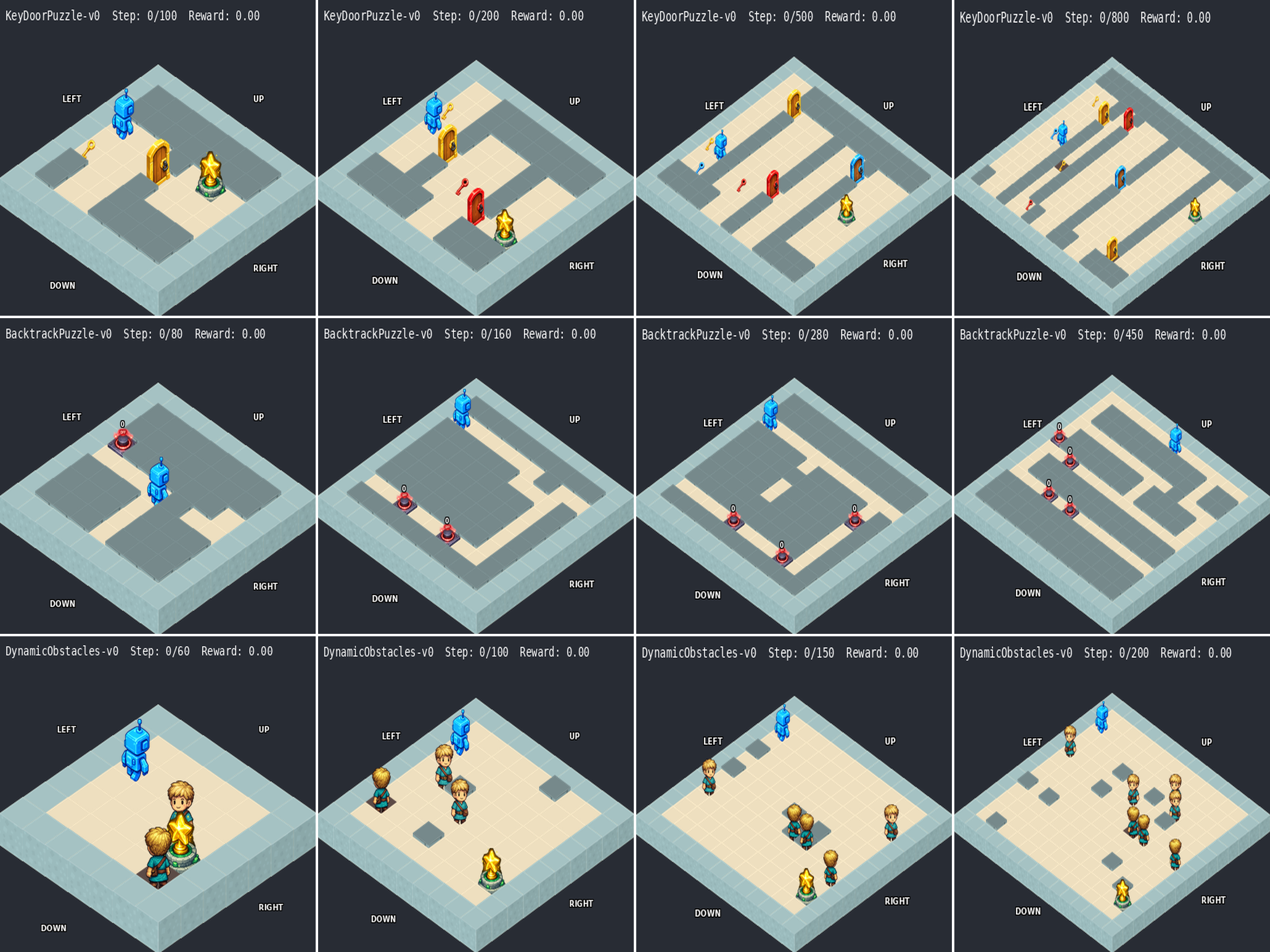}
    \caption{Three tasks at all four difficulty levels (columns: easy, medium, hard, expert). \textbf{Top:} KeyDoorPuzzle, where grid size, number of key-door pairs, and backtracking requirements increase. \textbf{Middle:} BacktrackPuzzle, with more switches, larger rooms, and tighter step budgets. \textbf{Bottom:} DynamicObstacles, with more obstacles, faster movement, and larger grids. Procedural generation ensures unique layouts at every seed.}
    \label{fig:difficulty_scaling}
\end{figure}

\revision{Difficulty labels select a task-specific generator configuration; they do not select hand-authored levels. For every seed, the generator samples a new instance while jointly scaling task-relevant axes such as grid size, object or constraint count, stochasticity, and horizon. The axes and four preset parameter ranges are benchmark design choices. Table~\ref{tab:difficulty_validation} empirically validates their intended ordering on four representative agents.}

\begin{table}[H]
  \caption{\revision{Capability-balanced mean success by difficulty over all 37 tasks and 25 official seeds per task--difficulty pair. Every evaluated regime degrades strictly with difficulty.}}
  \label{tab:difficulty_validation}
  \centering
  \small
  \setlength{\tabcolsep}{7pt}
  \revision{\begin{tabular}{lccccc}
    \agheader Agent & Easy & Medium & Hard & Expert & Easy$\rightarrow$Expert drop \\
    \agrowA GPT-5 mini & .494 & .332 & .223 & .189 & 61.8\% \\
    \agrowB PPO 2M & .606 & .277 & .160 & .104 & 82.8\% \\
    \agrowA Qwen3.5-4B & .445 & .259 & .110 & .098 & 78.0\% \\
    \agrowB SFT-250k & .647 & .467 & .369 & .303 & 53.1\% \\
    \midrule
    \agrowA Mean & \textbf{.548} & \textbf{.334} & \textbf{.215} & \textbf{.173} & \textbf{68.3\%} \\
  \end{tabular}}
\end{table}

\clearpage
\section{Full Task Descriptions}
\label{app:tasks}

\begin{figure}[H]
    \centering
    \includegraphics[width=\linewidth]{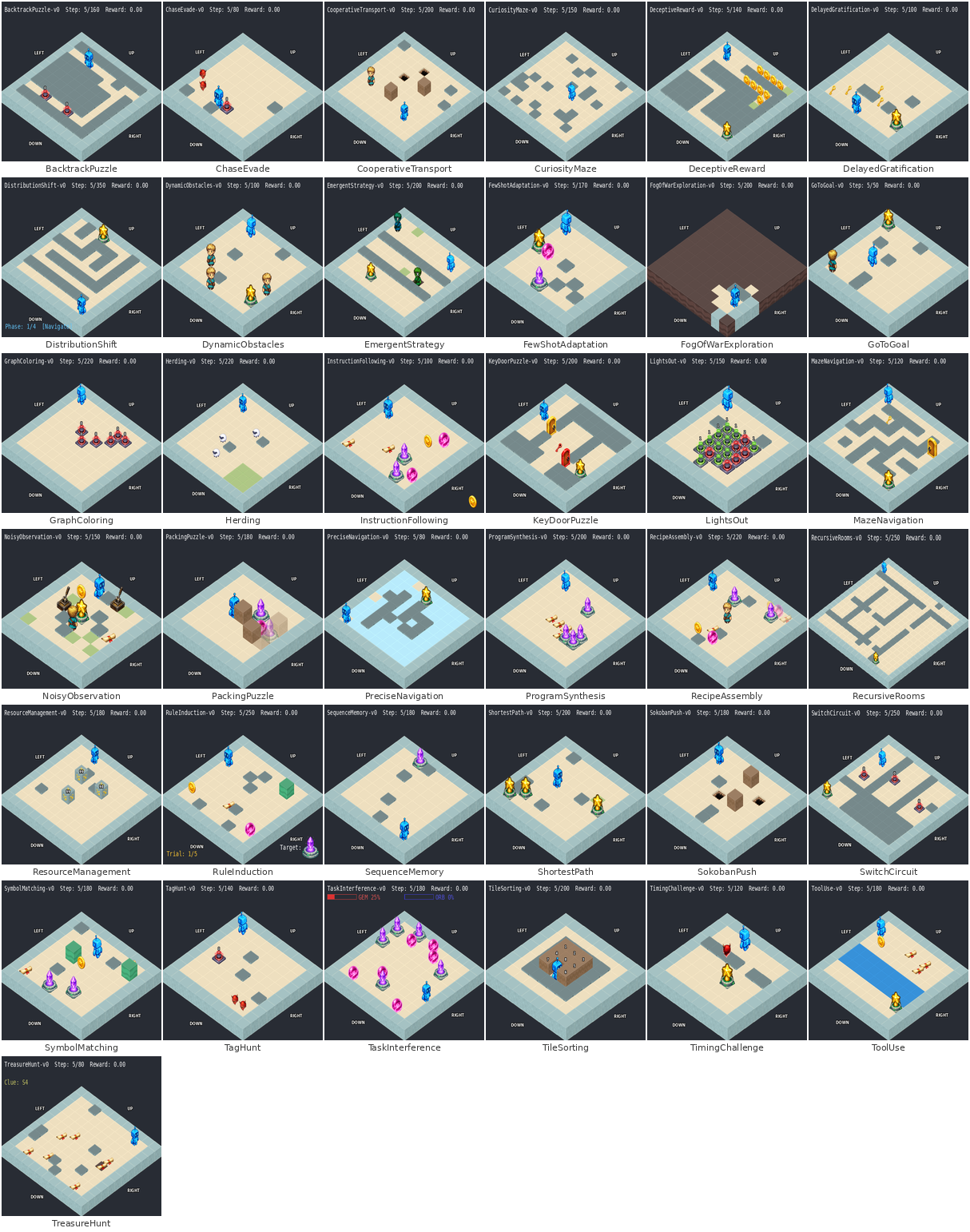}
    \caption{All 37 Agentick tasks in isometric view at medium difficulty, spanning six categories: navigation (8), planning (9), reasoning (8), memory (4), generalization (3), multi-agent (5).}
    \label{fig:task_gallery}
\end{figure}

Figure~\ref{fig:task_gallery} shows all 37 tasks rendered in isometric view at medium difficulty. Table~\ref{tab:full_tasks} provides descriptions and difficulty scaling.

\begin{table}[H]
  \caption{All 37 Agentick tasks with category, description, and difficulty scaling dimensions.}
  \label{tab:full_tasks}
  \centering
  \small
  \setlength{\tabcolsep}{3pt}
  \begin{tabular}{llp{5.5cm}p{3.5cm}}
    \agheader Category & Task & Description & Difficulty Scaling \\
    \agrowA \multirow{8}{*}{Navigation}
    & GoToGoal & Navigate to a visible goal & Grid size, obstacles \\
    \agrowB & MazeNavigation & Solve procedural mazes & Maze size, dead ends \\
    \agrowA & ShortestPath & Multi-goal TSP under step budget & \# goals, grid size \\
    \agrowB & DynamicObstacles & Reach goal dodging moving obstacles & \# obstacles, speed \\
    \agrowA & CuriosityMaze & Coverage-based exploration & Grid size, coverage \% \\
    \agrowB & RecursiveRooms & Hierarchical nested rooms & Nesting depth \\
    \agrowA & TimingChallenge & Cross patrol gaps precisely & \# patrols, speed \\
    \agrowB & InstructionFollowing & Find named target among distractors & \# distractors \\
    \midrule
    \agrowA \multirow{9}{*}{Planning}
    & SokobanPush & Push boxes onto targets & \# boxes, grid size \\
    \agrowB & KeyDoorPuzzle & Color-coded key-door chains & \# keys, backtracking \\
    \agrowA & BacktrackPuzzle & Activate switches, retrace path & \# switches, rooms \\
    \agrowB & TileSorting & Sliding tile puzzle & Tile count (8/15) \\
    \agrowA & PackingPuzzle & Push typed pieces to targets & \# pieces, types \\
    \agrowB & PreciseNavigation & Ice-sliding with wall segments & Grid size, stops \\
    \agrowA & RecipeAssembly & Ordered ingredient delivery & \# ingredients \\
    \agrowB & ToolUse & Discover scroll combinations & \# scrolls, combos \\
    \agrowA & ResourceManagement & Balance energy drain & \# stations, drain \\
    \midrule
    \agrowB \multirow{8}{*}{Reasoning}
    & SwitchCircuit & Non-linear switch dependencies & \# zones, exclusions \\
    \agrowA & RuleInduction & Discover hidden rules from cues & \# rules, wrong penalty \\
    \agrowB & LightsOut & Toggle lights with propagation & Grid size, toggles \\
    \agrowA & GraphColoring & Color nodes, no adjacent conflicts & \# nodes, chromatic \# \\
    \agrowB & SymbolMatching & Match and deliver symbol pairs & \# symbols, types \\
    \agrowA & ProgramSynthesis & Replicate reference pattern & \# gems, pattern size \\
    \agrowB & TaskInterference & Balance competing meters & Cross-drain, penalty \\
    \agrowA & DeceptiveReward & Resist misleading gradients & Decoy strength \\
    \midrule
    \agrowB \multirow{4}{*}{Memory}
    & SequenceMemory & Reproduce spatial sequences & Sequence length \\
    \agrowA & DelayedGratification & Resist traps for distant goal & \# decoys, distance \\
    \agrowB & TreasureHunt & Triangulate from scroll clues & \# clues, grid size \\
    \agrowA & FogOfWarExploration & Navigate with fog, 1-cell vis. & Grid size, obstacles \\
    \midrule
    \agrowB \multirow{3}{*}{General.}
    & FewShotAdaptation & Infer rule from demos & \# demos, rules \\
    \agrowA & DistributionShift & Multi-phase shifting maze & \# phases, remaps \\
    \agrowB & NoisyObservation & Navigate through noisy obs. & Noise level, ghosts \\
    \midrule
    \agrowA \multirow{5}{*}{Multi-Agent}
    & CooperativeTransport & Push heavy boxes cooperatively & \# boxes, \# partners \\
    \agrowB & TagHunt & Tag fleeing targets & \# targets, speed \\
    \agrowA & ChaseEvade & Survive pursuing adversaries & \# adversaries, types \\
    \agrowB & Herding & Guide sheep into pen & \# sheep, wander rate \\
    \agrowA & EmergentStrategy & Exploit entity behaviors & \# entities, behavior types \\
  \end{tabular}
\end{table}

\clearpage
\section{All Observation Modalities for a Single State}
\label{app:obs_modalities}

\begin{figure}[H]
\small

\textbf{(a) Isometric Pixel Rendering} (512$\times$512, uint8 RGB array; for VLM agents, CNN-RL, human play):

\begin{center}
\includegraphics[width=0.3\linewidth]{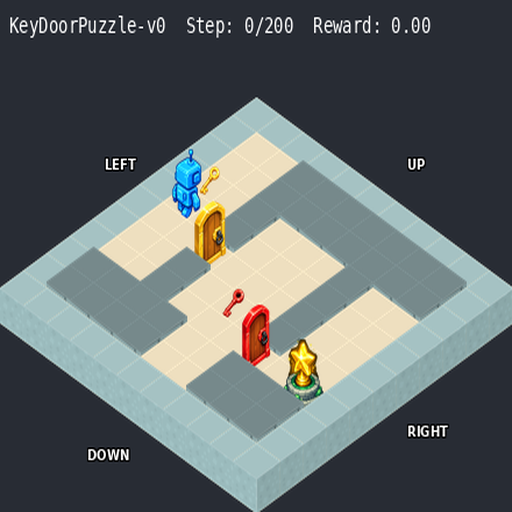}
\end{center}

\textbf{(b) ASCII} (colored text grid with legend; for LLM agents):
\begin{verbatim}
  # # # # # # # # # # #       Legend:
  # . . # # # # # # # #       ^ v < > : Agent (facing)
  # . . # # # # # # # #       # : Wall   . : Empty
  # Kg. # . . . # . . #       G : Goal   K : Key
  # ^ . # . . . # . . #       D : Door (closed)
  # . . Dg. . . # . . #       g=gold r=red
  # . . # . Kr. # . . #
  # . . # . . . Dr. G #
  # # # # # . . # # # #
  # # # # # . . # # # #
  # # # # # # # # # # #
\end{verbatim}

\textbf{(c) Natural Language} (spatial description; for LLM agents):

\begin{quote}
\textit{You are near the western edge of a 11$\times$11 room. You are facing north. You see: a gold key ahead (1 step), a closed gold door to your right (3 steps), a red key to your right (6 steps), a closed red door to your right (9 steps), a goal to your right (11 steps). Your inventory is empty. Actions: noop, move\_up, move\_down, move\_right, interact.}
\end{quote}

\textbf{(d) Structured Language} (parsed dictionary; for LLM/programmatic agents):
\begin{verbatim}
  {"description": "A 11x11 gridworld environment",
   "position": {"x": 1, "y": 4},
   "orientation": "north",
   "visible_entities": [
     {"type": "key", "position": [1,3], "distance": 1, "color": "gold"},
     {"type": "door", "position": [3,5], "distance": 3, "color": "gold"},
     {"type": "key", "position": [5,6], "distance": 6, "color": "red"},
     {"type": "door", "position": [7,7], "distance": 9, "color": "red"},
     {"type": "goal", "position": [9,7], "distance": 11}],
   "inventory": [],
   "valid_actions": ["noop","move_up","move_down","move_right","interact"],
   "step_count": 0, "max_steps": 200}
\end{verbatim}

\textbf{(e) State Dictionary} (raw numpy arrays; for RL and programmatic agents):
\begin{verbatim}
  grid.terrain:  int8[11,11]  (0=empty, 1=wall)
  grid.objects:  int8[11,11]  (0=none, 1=goal, 2=key, 3=door)
  grid.agents:   int8[11,11]  (agent position mask)
  grid.metadata: int16[11,11] (color/state encoding)
  agent.position:    (1, 4)
  agent.orientation: north
  agent.inventory:   []
  agent.energy:      1.0
\end{verbatim}

\caption{All five observation modalities for the same KeyDoorPuzzle state (medium, seed~42). Every modality is produced simultaneously from the same underlying grid, ensuring consistency across agent paradigms.}
\label{fig:all_obs}
\end{figure}

\section{Observation Modalities Summary}
\label{app:obs_modes_table}

\begin{table}[H]
  \caption{Observation modalities in Agentick. All modes are available for every task simultaneously.}
  \label{tab:obs_modes}
  \centering
  \small
  \begin{tabular}{llll}
    \agheader Mode & Format & Target Agents & Space \\
    \agrowA ASCII & Colored text grid + legend & LLM & String \\
    \agrowB Language & Natural language description & LLM & String \\
    \agrowA Language Structured & Dict (position, surroundings, actions) & LLM, Programmatic & Dict \\
    \agrowB Isometric Pixels & 512$\times$512 sprite rendering & VLM, CNN-RL, Human & Box \\
    \agrowA State Dict & Numpy arrays (terrain, objects, metadata) & RL, Programmatic & Dict \\
  \end{tabular}
\end{table}

\section{Per-Category ONS Results}
\label{app:results}

\revision{Table~\ref{tab:full_results} reports per-category ONS for representative complete configurations.}

\begin{table}[H]
  \caption{\revision{Per-category ONS for representative complete configurations. Best per category in \textbf{bold}. The rightmost column reports the 95\% bootstrap confidence interval on overall ONS; ``--'' marks pending estimates.}}
  \label{tab:full_results}
  \centering
  \small
  \setlength{\tabcolsep}{4pt}
  \begin{tabular}{lcccccc|cc}
    \agheader Agent & Nav. & Plan. & Reas. & Mem. & Gen. & Multi. & Overall & 95\% CI \\
    \agrowA \revision{SFT-250k} & \revision{\textbf{.568}} & \revision{\textbf{.557}} & \revision{\textbf{.445}} & \revision{\textbf{.413}} & \revision{.350} & \revision{.348} & \revision{\textbf{.447}} & \revision{{\scriptsize [.38,\,.52]}} \\
    \agrowB \revision{SFT-120k} & \revision{.428} & \revision{.459} & \revision{.350} & \revision{.288} & \revision{.357} & \revision{.246} & \revision{.354} & \revision{{\scriptsize [.29,\,.41]}} \\
    \agrowA GPT-5 mini & .456 & .334 & .131 & .348 & \textbf{.437} & .150 & .309 & {\scriptsize --} \\
    \agrowB PPO (2M) & .250 & .402 & .191 & .283 & .163 & \textbf{.432} & .287 & {\scriptsize [.21,\,.37]} \\
    \agrowA Qwen3.5-4B & .223 & .313 & .124 & .248 & .327 & .134 & .228 & {\scriptsize [.16,\,.29]} \\
    \agrowB PPO (500k) & .193 & .300 & .153 & .228 & .130 & .352 & .226 & {\scriptsize [.17,\,.29]} \\
    \agrowA Gemini 2.5 FL & .238 & .249 & .090 & .163 & .287 & .098 & .187 & {\scriptsize [.13,\,.25]} \\
    \agrowB Qwen3.5-2B & .136 & .237 & .048 & .213 & .133 & .032 & .133 & {\scriptsize [.07,\,.19]} \\
    \agrowA Qwen3.5-0.8B & .069 & .164 & .021 & .133 & .143 & .036 & .094 & {\scriptsize [.05,\,.14]} \\
    \agrowB Qwen3-4B & .073 & .106 & .005 & .153 & .133 & .038 & .085 & {\scriptsize --} \\
  \end{tabular}
\end{table}

\clearpage
\section{\revision{PPO Learning Curves}}
\label{app:ppo_curves}

\begin{figure}[p]
    \centering
    \includegraphics[width=\linewidth]{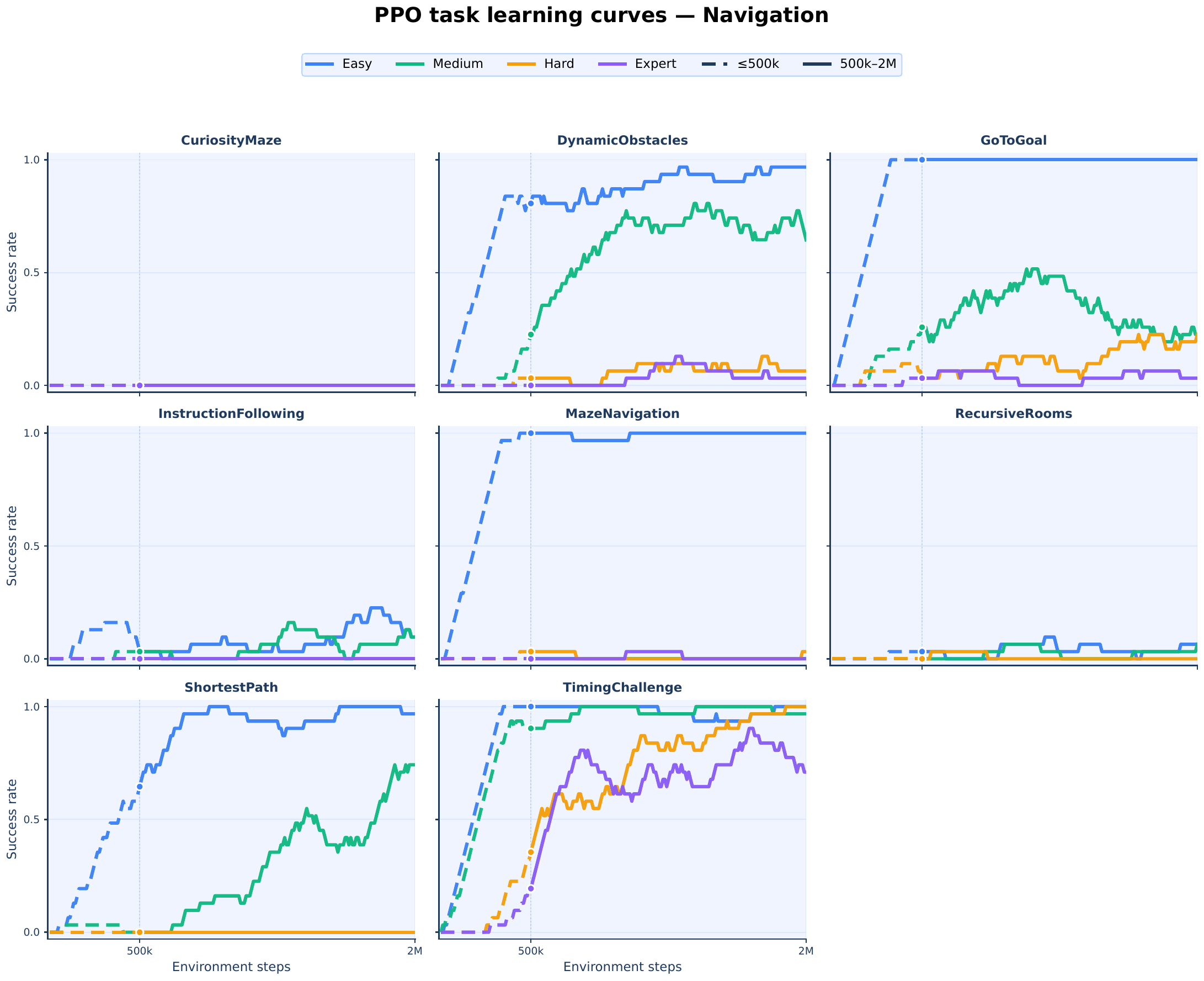}
    \caption{\revision{PPO learning curves for navigation. Each subplot is one task and each color is one independently trained difficulty-specific policy.}}
    \label{fig:ppo_curves_navigation}
\end{figure}

\begin{figure}[p]
    \centering
    \includegraphics[width=\linewidth]{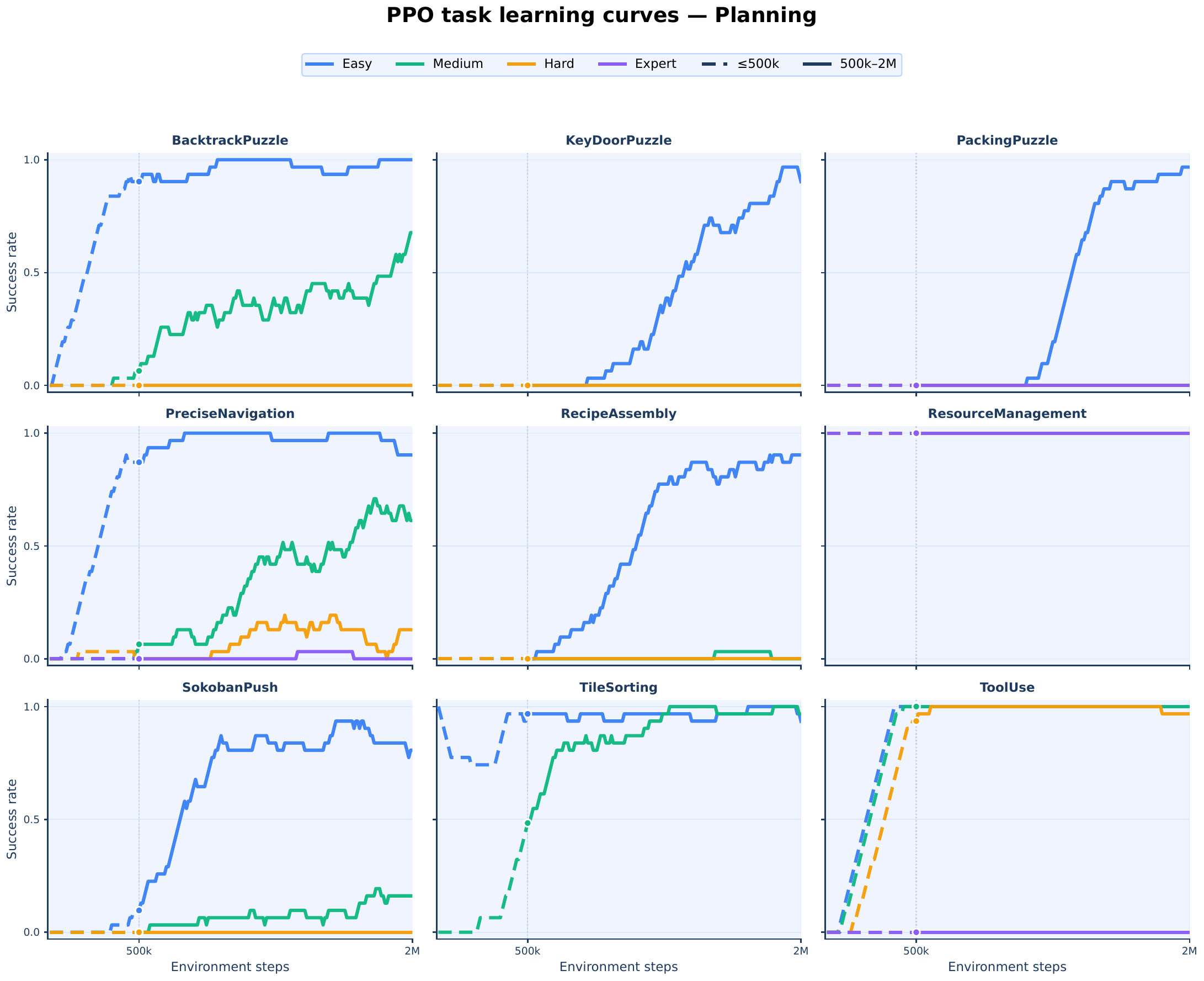}
    \caption{\revision{PPO learning curves for planning.}}
    \label{fig:ppo_curves_planning}
\end{figure}

\begin{figure}[p]
    \centering
    \includegraphics[width=\linewidth]{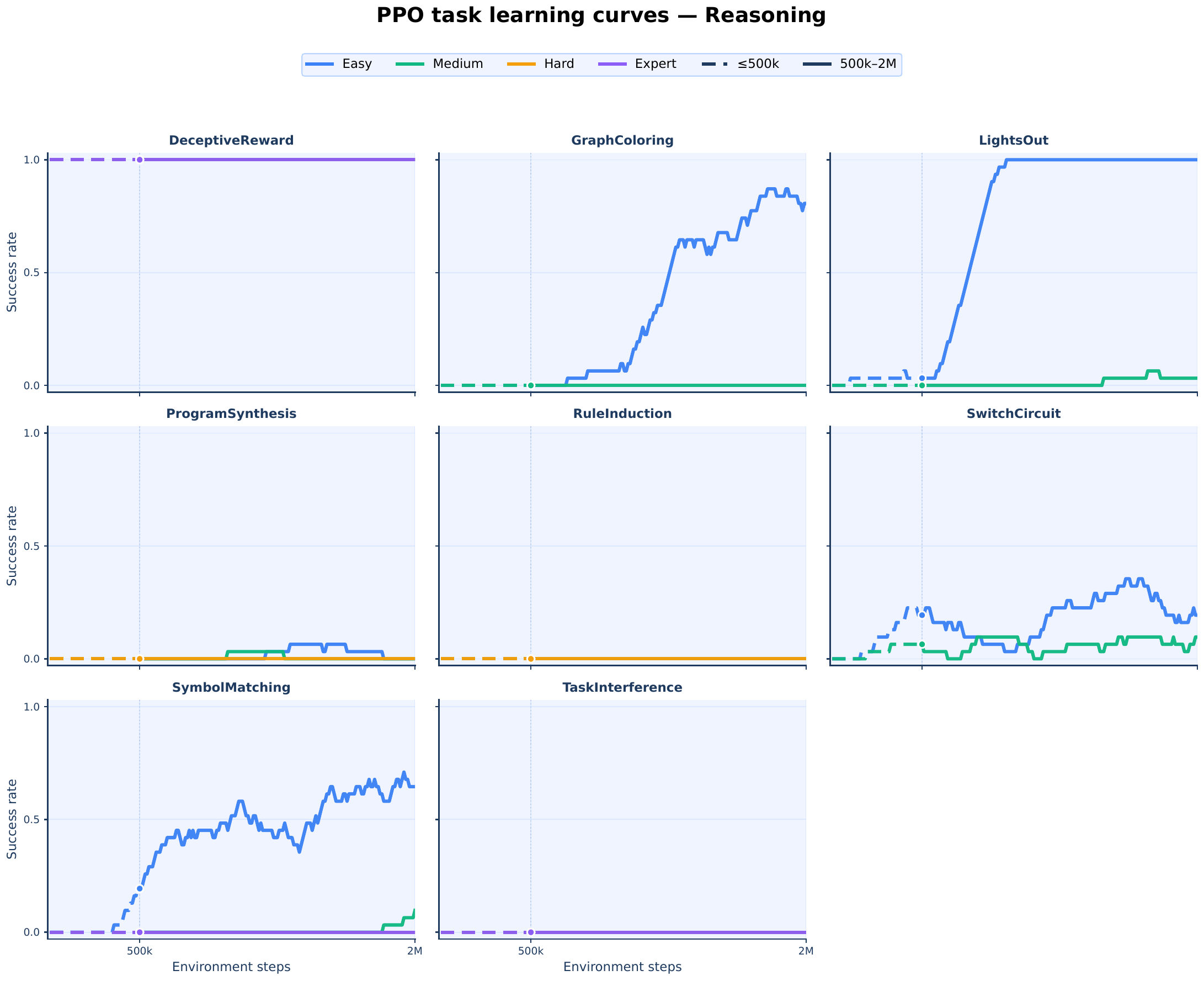}
    \caption{\revision{PPO learning curves for reasoning.}}
    \label{fig:ppo_curves_reasoning}
\end{figure}

\begin{figure}[p]
    \centering
    \includegraphics[width=\linewidth]{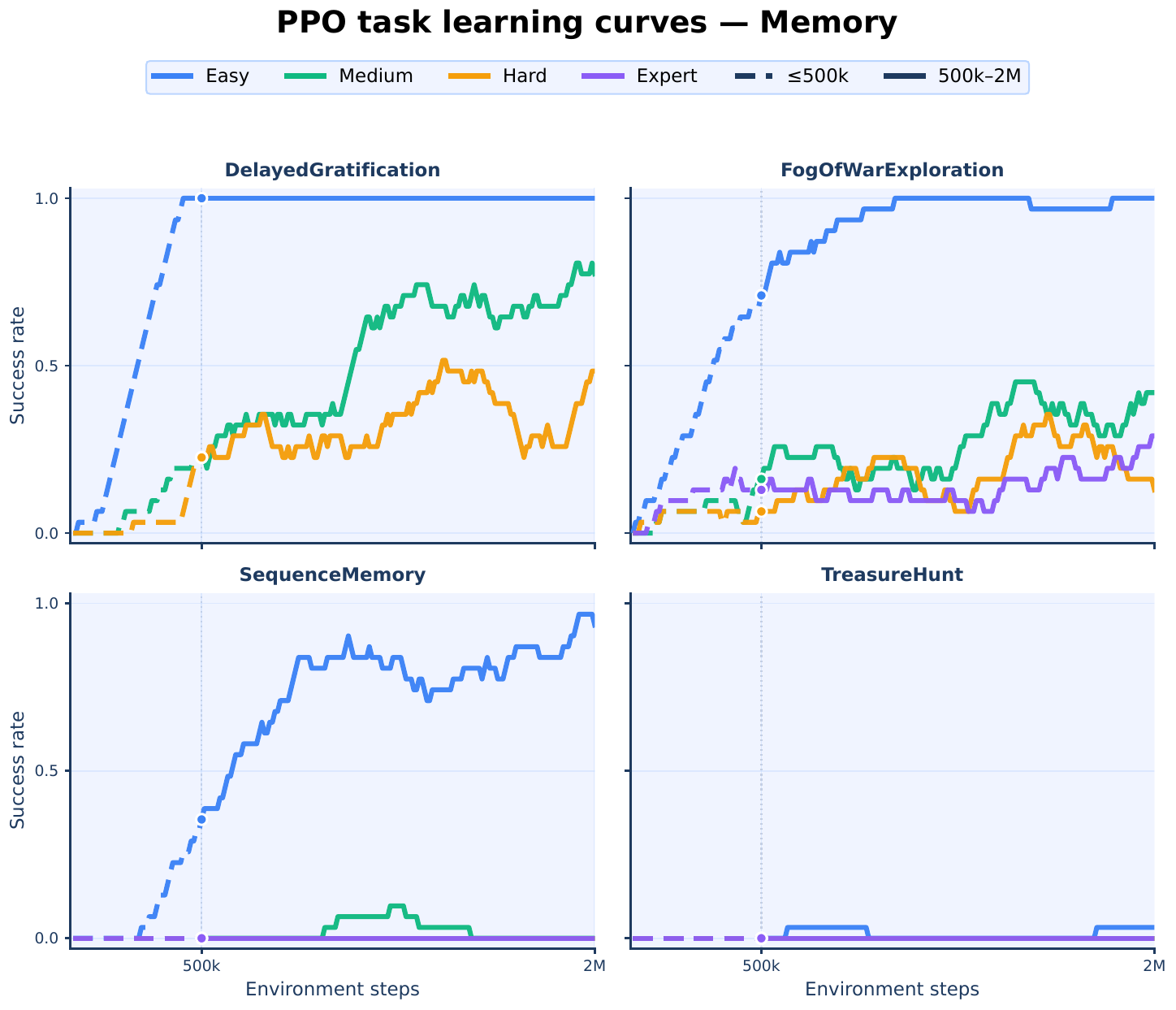}
    \caption{\revision{PPO learning curves for memory.}}
    \label{fig:ppo_curves_memory}
\end{figure}

\begin{figure}[p]
    \centering
    \includegraphics[width=\linewidth]{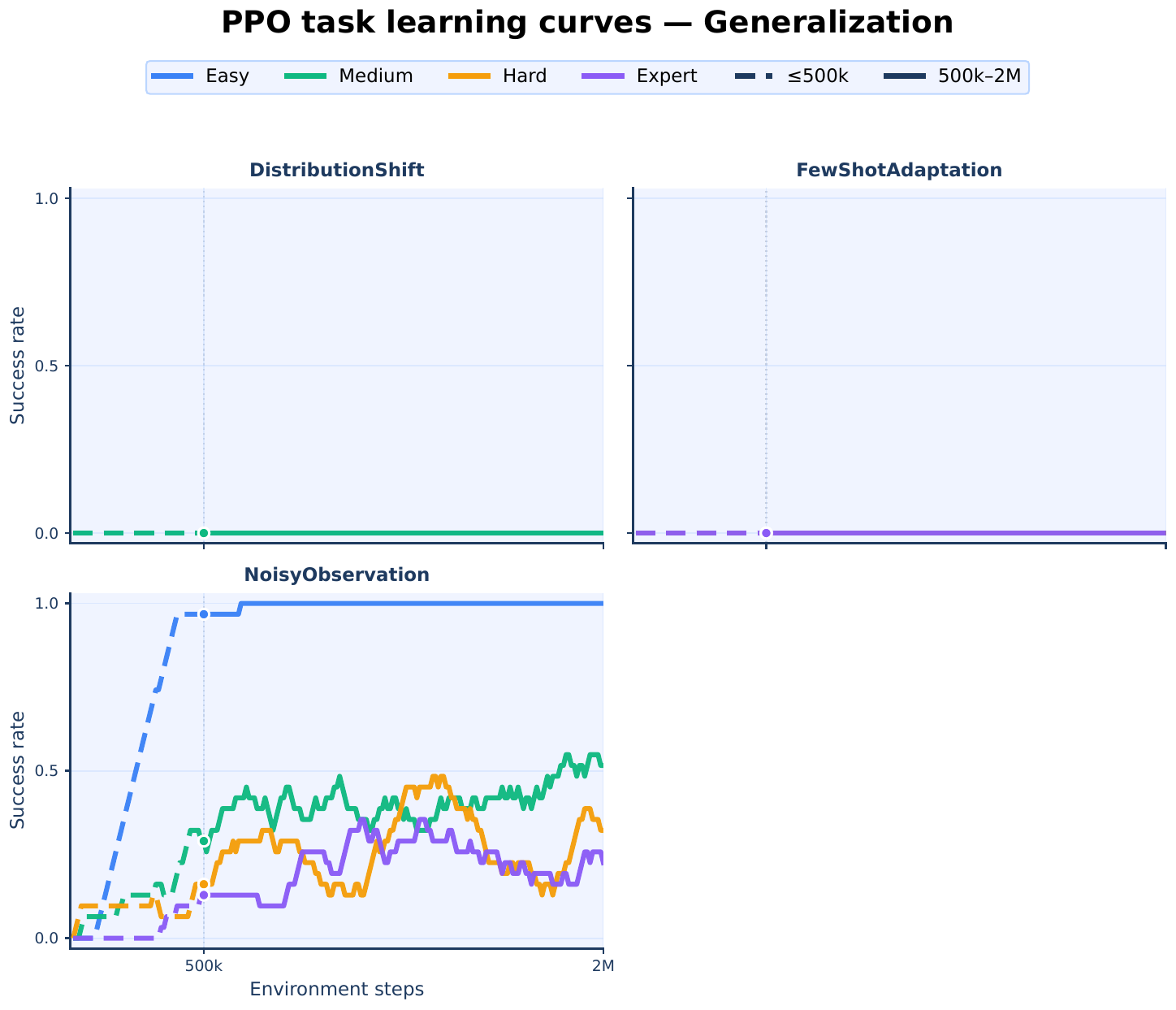}
    \caption{\revision{PPO learning curves for generalization.}}
    \label{fig:ppo_curves_generalization}
\end{figure}

\begin{figure}[p]
    \centering
    \includegraphics[width=\linewidth]{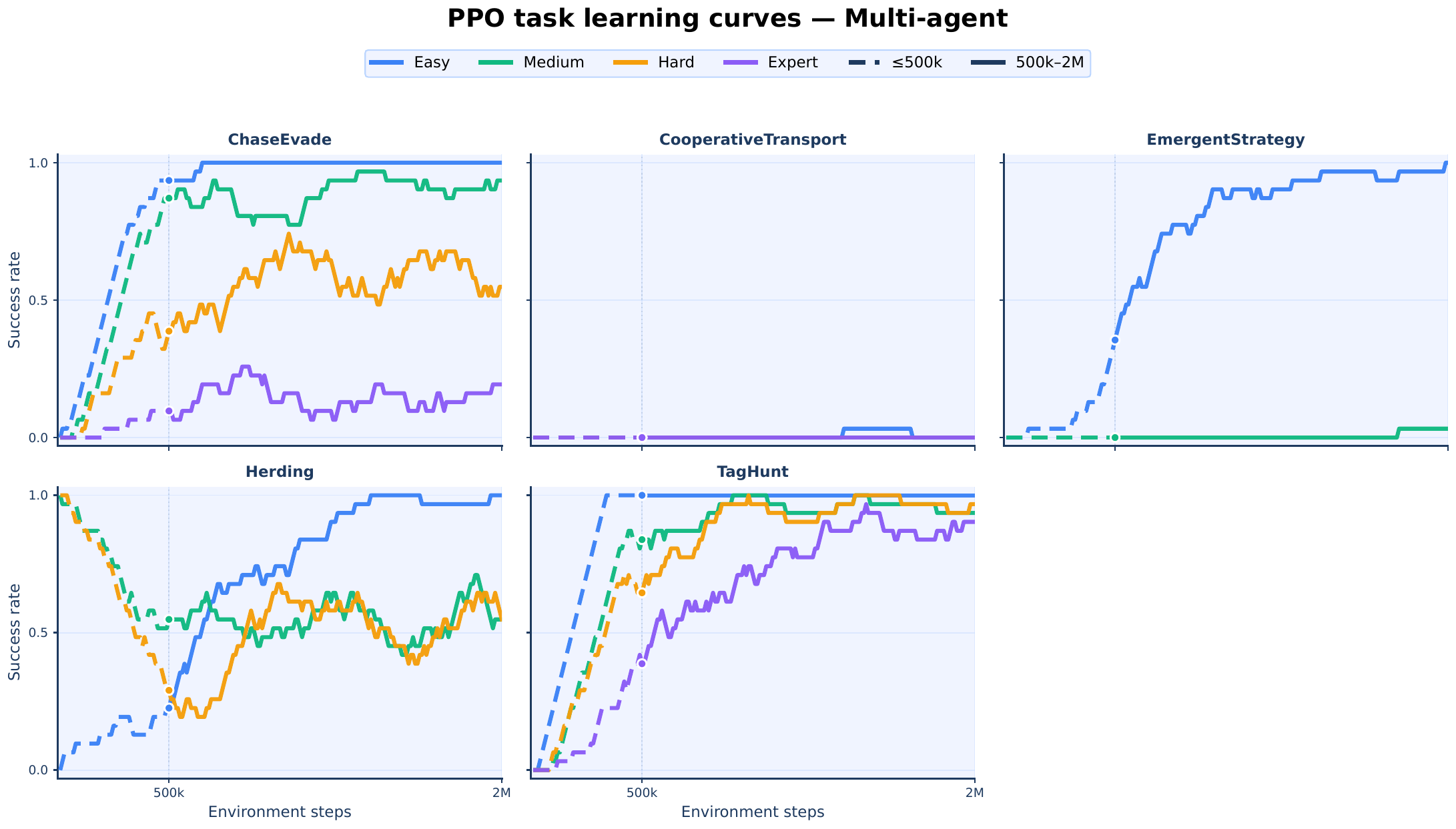}
    \caption{\revision{PPO learning curves for multi-agent tasks.}}
    \label{fig:ppo_curves_multiagent}
\end{figure}

\revision{Figures~\ref{fig:ppo_curves_navigation}--\ref{fig:ppo_curves_multiagent} show the full recovered 2M-step learning curves for 126 of the 148 independently trained task--difficulty policies. Color denotes difficulty. Each curve is dashed through 500k steps, marked at 500k, and solid thereafter; values are periodic evaluation success smoothed with a 31-point trailing mean. These diagnostic evaluations use one deterministic episode per point, whereas the reported final metrics use 25 official evaluation seeds.}

\begin{table}[H]
  \caption{\revision{Smoothed periodic PPO success at fixed checkpoints. The task-balanced gain falls to $+0.016$ in the final 500k steps; aggregate success preserves easy $>$ medium $>$ hard $>$ expert at every checkpoint.}}
  \label{tab:ppo_curve_checkpoints}
  \centering
  \small
  \setlength{\tabcolsep}{7pt}
  \revision{\begin{tabular}{lccccc}
    \agheader Recovered policies & 500k & 1M & 1.5M & 2M & $\Delta$ 1.5M$\rightarrow$2M \\
    \agrowA Easy (37) & .441 & .573 & .656 & .679 & +.024 \\
    \agrowB Medium (37) & .217 & .282 & .313 & .339 & +.026 \\
    \agrowA Hard (31) & .167 & .221 & .243 & .239 & $-.003$ \\
    \agrowB Expert (21) & .141 & .181 & .209 & .209 & +.000 \\
    \midrule
    \agrowA Task-balanced mean & \textbf{.243} & \textbf{.329} & \textbf{.377} & \textbf{.393} & \textbf{+.016} \\
  \end{tabular}}
\end{table}

\revision{From 1.5M to 2M, the median task-level absolute change is .016; 32/37 task aggregates change by at most .05 and 36/37 by at most .10. At 2M, 17/21 tasks with all four recovered curves order easy $\geq$ medium $\geq$ hard $\geq$ expert.}

\section{Agent Configuration Details}
\label{app:configs}

\noindent \textbf{Frontier LLMs.} GPT-5~mini, Gemini~3.1~Flash~Lite, and Claude~Haiku~4.5 were evaluated via their respective APIs with temperature~0, max tokens~100, using ASCII observations and the Markovian Reasoner harness. Each model was evaluated on all 37 tasks $\times$ 4 difficulties $\times$ 25 seeds.

\noindent \textbf{PPO.} Trained using Stable-Baselines3~\citep{raffin2021stable} with CnnPolicy on pixel observations preprocessed in the ALE style: 512$\times$512 RGB isometric renders were resized to 84$\times$84, converted to grayscale, and stacked over four frames. \revision{We train one policy independently for each of the 37 tasks and four difficulty levels; PPO is not multitask.} We used $n_\text{steps}=128$, batch size~256, learning rate $2.5 \times 10^{-4}$, dense reward mode, and 2M total timesteps for the headline PPO baseline. Training was performed on a single NVIDIA A100 GPU.

\noindent \textbf{Qwen models.} All Qwen models (Qwen3-4B, Qwen3.5-0.8B, Qwen3.5-2B, Qwen3.5-4B) were served via vLLM with temperature~0.7, top-$p$~0.8, top-$k$~20, max tokens~100. Each model was evaluated across all 4 combinations of \{ASCII, language\} $\times$ \{Markovian, Markovian Reasoner\}. The best configuration per model is reported in the main text.

\revision{\noindent \textbf{SFT.} We LoRA-fine-tuned Qwen3.5-4B for two epochs on either 120k or 250k per-step examples collected from Agentick's deterministic oracle policies. Each example pairs an ASCII state with the oracle action and contains no chain-of-thought rationale. We evaluated each checkpoint under the same Markovian and Markovian Reasoner harnesses and Qwen sampling settings as the base model, across all 37 tasks, four difficulty levels, and 25 official seeds.}

\revision{\noindent \textbf{LLM system prompt.} At every step, both harnesses receive the same task-specific system template (with the task description substituted in braces): \textit{``You are an AI agent playing grid-world tasks in the Agentick benchmark. Your goal is to navigate the grid and complete the task objective by selecting the best action at each step. Action space: 0 NOOP; 1 MOVE\_UP; 2 MOVE\_DOWN; 3 MOVE\_LEFT; 4 MOVE\_RIGHT; 5 INTERACT (use the object in the cell you are facing; for solid switches, levers, or doors, stand adjacent and face the object first). Task objective: \{task description\}. Instructions: (1) Observe the current grid state. (2) Reason about the best action to take. (3) Output your selected action as a single integer (0--5). Respond with ONLY the action number, nothing else.''} The user message contains only the current observation; neither harness receives trajectory history.}

\noindent \textbf{Reasoner harness prompt.} The Reasoner harness appends the following to the system prompt: \textit{``IMPORTANT: Before choosing an action, reason step-by-step but be CONCISE (2--4 sentences max): 1.~What do you observe? What is your goal? 2.~Which action best advances you toward the goal? 3.~Output your final answer on the LAST line as: ACTION: $\langle$number$\rangle$.''}

\section{Compute Budget}
\label{app:compute}

\noindent \textbf{RL training.} PPO training for 2M steps required approximately 4 GPU-hours on a single NVIDIA A100 (80GB) per task. Total RL compute: $\sim$150 GPU-hours across all 37 tasks.

\noindent \textbf{LLM evaluation.} Frontier model evaluation used commercial APIs. Approximate costs: GPT-5~mini \$180, Gemini~3.1~Flash~Lite \$45, Claude~Haiku~4.5 \$65. Open-weight Qwen models were served on 2$\times$A100 GPUs via vLLM; total serving time $\sim$200 GPU-hours across all model sizes and configurations.

\noindent \textbf{Oracle trajectory generation.} The 500K-episode dataset was generated in $\sim$8 CPU-hours on a 32-core machine.

\clearpage
\section{Coding API}
\label{app:coding_api}

Every Agentick environment exposes a programmatic \textbf{Coding API} (\texttt{AgentickAPI}) that provides spatial queries, BFS pathfinding, entity lookups, and high-level action primitives. The API is designed for three use cases: (1)~writing hand-coded bot agents and planners, (2)~building oracle policies (all 37 oracles in Agentick are implemented through this API), and (3)~enabling code-generating LLMs to write agent logic in Python rather than selecting raw action integers.

\noindent \textbf{Setup.} The API wraps an environment and is updated after each step:
\begin{verbatim}
  from agentick.coding_api import AgentickAPI
  import agentick

  env = agentick.make("KeyDoorPuzzle-v0", difficulty="medium")
  api = AgentickAPI(env)
  obs, info = env.reset(seed=42)
  api.update(obs, info)
\end{verbatim}

\noindent \textbf{Spatial queries.} The API exposes the agent's position, orientation, and spatial relationships to all entities:
\begin{verbatim}
  api.agent_position             # (1, 4)
  api.agent_direction            # 'north'
  api.grid_size                  # (11, 11)
  api.get_nearest("key")         # EntityInfo(type='key', pos=(1,3), dist=1)
  api.get_nearest("goal")        # EntityInfo(type='goal', pos=(9,7), dist=11)
  api.get_entities_of_type("key")# [EntityInfo(...), EntityInfo(...)]
  api.get_entity_at(3, 5)        # EntityInfo(type='door', ...)
  api.is_walkable(2, 4)          # True
  api.is_walkable(3, 4)          # False (wall)
  api.is_reachable(9, 7)         # False (blocked by locked doors)
  api.distance_to(1, 3)          # 1
  api.direction_to(1, 3)         # 'north'
  api.is_adjacent(1, 3)          # True
\end{verbatim}

\noindent \textbf{BFS pathfinding.} The API provides shortest-path computation that respects walls, blocking objects, and terrain:
\begin{verbatim}
  api.path_to(1, 3)              # [1]  (action sequence to gold key)
  api.go_to_nearest("key")       # [1]  (pathfind to closest key)
  api.go_to_nearest("door")      # [1, 1, 3, 3, 3] (to gold door)
  api.flee_from(5, 5)            # single action moving away from (5,5)
  api.move_toward(9, 7)          # single action toward goal
\end{verbatim}
The \texttt{path\_to} and \texttt{go\_to\_nearest} methods return sequences of action integers that can be executed directly via \texttt{env.step()}. The pathfinder accounts for non-walkable objects (locked doors, walls) and updates dynamically as the environment state changes.

\noindent \textbf{Grid inspection.} Low-level access to the grid structure:
\begin{verbatim}
  api.get_walkable_cells()       # [(1,1), (1,2), ...] (43 cells)
  api.get_walls()                # [(0,0), (0,1), ...] (all wall positions)
  api.get_cell(3, 5)             # {'terrain': 'empty', 'object': 'door'}
  api.get_object(1, 3)           # 'key'
  api.get_terrain_type(0, 0)     # 'wall'
  api.neighbors(1, 4)            # [(1,3), (2,4), (1,5)] (walkable neighbors)
\end{verbatim}

\noindent \textbf{Inventory and interaction.} For tasks involving item collection and object interaction:
\begin{verbatim}
  api.get_inventory()            # []
  api.has_in_inventory("key")    # False
  api.interact_with(1, 3)        # action sequence: face + INTERACT
  api.pickup_nearest("key")      # pathfind + interact
\end{verbatim}

\noindent \textbf{Execution helpers.} High-level methods for stepping through action sequences:
\begin{verbatim}
  api.step_action(1)             # execute action 1 (move_up), return obs
  api.valid_actions              # [0, 1, 2, 4, 5]
  api.action_names               # ['noop','move_up','move_down',...]
  api.action_name_to_int("move_up")  # 1
  api.current_step               # 0
  api.max_steps                  # 200
  api.total_reward               # 0.0
  api.is_done                    # False
\end{verbatim}

\noindent \textbf{Oracle construction.} All 37 oracle policies are implemented using this API. A typical oracle follows the pattern:
\begin{verbatim}
  class KeyDoorOracle(OracleAgent):
      def plan(self):
          # Find nearest uncollected key
          key = self.api.get_nearest("key")
          if key:
              self.action_queue = self.api.go_to_nearest("key")
          else:
              # All keys collected, head to goal
              self.action_queue = self.api.go_to_nearest("goal")

      def act(self, obs, info):
          self.api.update(obs, info)
          if not self.action_queue:
              self.plan()
          return self.action_queue.pop(0)
\end{verbatim}
This API-based oracle design means that expert trajectories are generated through interpretable, verifiable code rather than opaque neural network policies, enabling researchers to inspect, debug, and modify the strategies that produce the SFT training data.

\clearpage   
\section*{NeurIPS Paper Checklist}

\begin{enumerate}

\item {\bf Claims}
    \item[] Question: Do the main claims made in the abstract and introduction accurately reflect the paper's contributions and scope?
    \item[] Answer: \answerYes{}
    \item[] Justification: The abstract and Section~\ref{sec:introduction} state the four contributions (37 procedurally generated tasks, capability decomposition, training-first infrastructure, and the empirical evaluation across 27 configurations and over 90,000 episodes) that are substantiated in Sections~\ref{sec:benchmark} and~\ref{sec:experiments}.

\item {\bf Limitations}
    \item[] Question: Does the paper discuss the limitations of the work performed by the authors?
    \item[] Answer: \answerYes{}
    \item[] Justification: Limitations are summarized in Section~\ref{sec:discussion} and detailed in Appendix~\ref{app:limitations}, covering the gridworld abstraction, capability coverage, observation equivalence, RL training budget, multi-agent scope, and evaluation coverage.

\item {\bf Theory assumptions and proofs}
    \item[] Question: For each theoretical result, does the paper provide the full set of assumptions and a complete (and correct) proof?
    \item[] Answer: \answerNA{}
    \item[] Justification: The paper presents an empirical benchmark and contains no theoretical results requiring proofs.

    \item {\bf Experimental result reproducibility}
    \item[] Question: Does the paper fully disclose all the information needed to reproduce the main experimental results of the paper to the extent that it affects the main claims and/or conclusions of the paper (regardless of whether the code and data are provided or not)?
    \item[] Answer: \answerYes{}
    \item[] Justification: Section~\ref{sec:evaluation} specifies deterministic seeds and the YAML-based runner; Appendix~\ref{app:configs} reports all agent configurations, hyperparameters, and the Reasoner harness prompt; and the public repository, datasets, and leaderboard provide the code and seeds needed to reproduce results.

\item {\bf Open access to data and code}
    \item[] Question: Does the paper provide open access to the data and code, with sufficient instructions to faithfully reproduce the main experimental results, as described in supplemental material?
    \item[] Answer: \answerYes{}
    \item[] Justification: The benchmark code, oracle policies, pre-built SFT datasets (120K/250K/500K episodes on HuggingFace), and live leaderboard are publicly available through the links listed below the abstract, and Section~\ref{sec:infrastructure} and Appendix~\ref{app:configs} provide the YAML-based experiment runner and configuration details.

\item {\bf Experimental setting/details}
    \item[] Question: Does the paper specify all the training and test details (e.g., data splits, hyperparameters, how they were chosen, type of optimizer) necessary to understand the results?
    \item[] Answer: \answerYes{}
    \item[] Justification: Section~\ref{sec:setup} describes agents, observation modes, harnesses, and the 25-seed evaluation protocol, and Appendix~\ref{app:configs} reports the PPO hyperparameters (learning rate, batch size, $n_\text{steps}$, dense reward, 2M timesteps), Qwen sampling settings, frontier-LLM API settings, and the exact Reasoner harness prompt.

\item {\bf Experiment statistical significance}
    \item[] Question: Does the paper report error bars suitably and correctly defined or other appropriate information about the statistical significance of the experiments?
    \item[] Answer: \answerYes{}
    \item[] Justification: Sections~\ref{sec:evaluation} and~\ref{sec:setup} report 95\% bootstrap confidence intervals computed over the 25 evaluation episodes per task--difficulty pair following~\citet{agarwal2021deep}; intervals are shown in figures and in Appendix~\ref{app:results} (Table~\ref{tab:full_results}).

\item {\bf Experiments compute resources}
    \item[] Question: For each experiment, does the paper provide sufficient information on the computer resources (type of compute workers, memory, time of execution) needed to reproduce the experiments?
    \item[] Answer: \answerYes{}
    \item[] Justification: Appendix~\ref{app:compute} reports PPO training cost ($\sim$4 GPU-hours per task on a single A100, $\sim$150 GPU-hours total), frontier-LLM API costs, vLLM serving cost for Qwen models ($\sim$200 GPU-hours on 2$\times$A100), and oracle dataset generation cost.
    
\item {\bf Code of ethics}
    \item[] Question: Does the research conducted in the paper conform, in every respect, with the NeurIPS Code of Ethics \url{https://neurips.cc/public/EthicsGuidelines}?
    \item[] Answer: \answerYes{}
    \item[] Justification: The work uses synthetic gridworld environments with no human subjects or scraped data, releases code and datasets publicly, and conforms with the NeurIPS Code of Ethics.

\item {\bf Broader impacts}
    \item[] Question: Does the paper discuss both potential positive societal impacts and negative societal impacts of the work performed?
    \item[] Answer: \answerYes{}
    \item[] Justification: Section~\ref{sec:discussion} discusses positive impacts (assistive AI, scientific discovery) and the risk of accelerating autonomous systems without adequate oversight, and encourages safety- and alignment-aware development.
    
\item {\bf Safeguards}
    \item[] Question: Does the paper describe safeguards that have been put in place for responsible release of data or models that have a high risk for misuse (e.g., pre-trained language models, image generators, or scraped datasets)?
    \item[] Answer: \answerNA{}
    \item[] Justification: The released artifacts are synthetic gridworld environments, oracle trajectories, and benchmark code; no pre-trained generative models, scraped data, or other high-risk assets are released.

\item {\bf Licenses for existing assets}
    \item[] Question: Are the creators or original owners of assets (e.g., code, data, models), used in the paper, properly credited and are the license and terms of use explicitly mentioned and properly respected?
    \item[] Answer: \answerYes{}
    \item[] Justification: All external assets are cited in the text, including Gymnasium~\citep{towers2024gymnasium}, Stable-Baselines3~\citep{raffin2021stable}, the Qwen models~\citep{qwen3technicalreport}, and the frontier model APIs; their licenses and terms of use are respected and listed in the supplementary repository.

\item {\bf New assets}
    \item[] Question: Are new assets introduced in the paper well documented and is the documentation provided alongside the assets?
    \item[] Answer: \answerYes{}
    \item[] Justification: The benchmark code, Coding API (Appendix~\ref{app:coding_api}), oracle policies, harness infrastructure, and SFT datasets (with Croissant metadata including Responsible-AI fields) are released publicly with documentation, a public docs site, and the leaderboard linked under the abstract.

\item {\bf Crowdsourcing and research with human subjects}
    \item[] Question: For crowdsourcing experiments and research with human subjects, does the paper include the full text of instructions given to participants and screenshots, if applicable, as well as details about compensation (if any)?
    \item[] Answer: \answerNA{}
    \item[] Justification: The paper does not involve crowdsourcing or research with human subjects.

\item {\bf Institutional review board (IRB) approvals or equivalent for research with human subjects}
    \item[] Question: Does the paper describe potential risks incurred by study participants, whether such risks were disclosed to the subjects, and whether Institutional Review Board (IRB) approvals (or an equivalent approval/review based on the requirements of your country or institution) were obtained?
    \item[] Answer: \answerNA{}
    \item[] Justification: The paper does not involve research with human subjects, so IRB approval is not applicable.

\item {\bf Declaration of LLM usage}
    \item[] Question: Does the paper describe the usage of LLMs if it is an important, original, or non-standard component of the core methods in this research? Note that if the LLM is used only for writing, editing, or formatting purposes and does \emph{not} impact the core methodology, scientific rigor, or originality of the research, declaration is not required.
    \item[] Answer: \answerNA{}
    \item[] Justification: LLMs were used only for grammar correction during paper writing and do not affect the core methodology, scientific rigor, or originality of the research.

\end{enumerate}

\end{document}